\documentclass[preprint,12pt,review]{elsarticle}
\usepackage{siunitx}
\usepackage{booktabs}
\usepackage{graphicx}
\usepackage{pgfplots}
\usepackage{pdflscape}
\usepackage[caption=false, subrefformat=parens]{subfig}
\pgfplotsset{compat=newest}
\usetikzlibrary{plotmarks}
\usetikzlibrary{arrows.meta}
\usepgfplotslibrary{patchplots}
\newlength\fwidth
\pgfplotsset{plot coordinates/math parser=false}
\usepackage{overpic}
\pgfplotsset{compat=1.14}
\usepackage{float}
\usepackage{etoolbox}
\usepackage{bicaption}
\usepackage{makecell}

\usepackage[skip=1ex, labelfont=bf]{caption}
\usepackage[skip=0.333\baselineskip]{caption}
\usepackage{tabularx}
\newcolumntype{C}{>{\centering\arraybackslash}X}

\usepackage[margin=2.5cm]{geometry} 
\usepackage[skip=0.333\baselineskip]{caption}
\usepackage{amssymb,amsmath}
\usepackage{url}
\usepackage{hyperref}
\usepackage{multirow}
\usepackage{arydshln}

\journal{Pattern Recognition}
\biboptions{sort&compress}
\begin{document}

\begin{frontmatter}

\title{TSAR-MVS: Textureless-aware Segmentation and Correlative Refinement Guided Multi-View Stereo}


    

\author[label1]{Zhenlong Yuan}
\author[label1]{Jiakai Cao}
\author[label1]{Zhaoqi Wang}
\author[label2,label3]{Zhaoxin Li\corref{cor1}}
\cortext[cor1]{Corresponding Author.\ead{cszli@hotmail.com}}

\affiliation[label1]{organization={Institute of Computing Technology, Chinese Academy of Sciences},
            city={Beijing},
            postcode={100190},
            country={China}
}

\affiliation[label2]{organization={Agricultural Information Institute, Chinese Academy of Agricultural Sciences},
            city={Beijing},
            postcode={100081},
            country={China}
}

\affiliation[label3]{organization={Key Laboratory of Agricultural Big Data, Ministry of Agriculture and Rural Affairs},
            city={Beijing},
            postcode={100081},
            country={China}
}






\begin{abstract}
The reconstruction of textureless areas has long been a challenging problem in MVS due to lack of reliable pixel correspondences between images. In this paper, we propose the Textureless-aware Segmentation And Correlative Refinement guided Multi-View Stereo (TSAR-MVS), a novel method that effectively tackles challenges posed by textureless areas in 3D reconstruction through filtering, refinement and segmentation. First, we implement the joint hypothesis filtering, a technique that merges a confidence estimator with a disparity discontinuity detector to eliminate incorrect depth estimations. Second, to spread the pixels with confident depth, we introduce an iterative correlation refinement strategy that leverages RANSAC to generate 3D planes based on superpixels, succeeded by a weighted median filter for broadening the influence of accurately determined pixels. 
Finally, we present a textureless-aware segmentation method that leverages edge detection and line detection for accurately identify large textureless regions for further depth completion.
Experiments on ETH3D, Tanks \& Temples and Strecha datasets demonstrate the superior performance and strong generalization capability of our proposed method. 
\end{abstract}



\end{frontmatter}

\section{Introduction}
Multi-view Stereo (MVS) aims to rebuild dense 3D scenes or objects from a series of calibrated RGB images. It has a broad range of applications in fields such as virtual reality, 3D printing, and autonomous driving, leading to numerous efficient MVS frameworks~\cite{LTVRE, Accurate, Depthmap, PMVS}. Nonetheless, despite the rapidly improving accuracy of the latest methods ~\cite{MVSNet, P-MVS}, these techniques frequently face challenges when dealing with textureless areas.

Recently, PatchMatch Stereo based-methods~\cite{PM-Huber, PCF-MVS} have accomplished remarkable successes in reconstructing dense 3D models from large-scale imagery. 
These methods typically follow an organized pipeline that includes initialization, propagation and refinement, positioning them uniquely suited for reconstructing scenes with unstructured viewpoints. 
However, due to the unreliable photometric consistency, most methods~\cite{COLMAP, Gipuma, MAR-MVS, SD-MVS} still exhibit unsatisfactory performance when dealing with large textureless areas.

Some traditional methods ~\cite{ACMM, ACMP, ACMMP} leverage multi-scale geometric consistency and triangulated planar prior to address this challenge. Through obtaining planar hypotheses at coarser scales and propagating reliable ones into finer scales, these methods manage to effectively reconstruct partial textureless areas. Yet, the integration of larger textureless areas remains problematic due to their limited patch sizes. Partial approaches~\cite{MESH, TAPA-MVS, PCF-MVS} attempt to utilize superpixels for 3D planar surface fitting to alleviate the problem. However, constraints such as a fixed threshold and strict assumption of data distribution result in poor planarization for certain superpixels. APD-MVS~\cite{APD-MVS} utilizes an adaptive deformable patch with cascade architecture. Nevertheless, excessive PatchMatch iterations often make the process time-consuming, leading it impractical for application on large-scale datasets.

Some learning-based methods, such as ~\cite{Cas-MVSNet, P-MVS} leverage a coarse-to-fine cascading architecture to address the issue of textureless areas. By refining and upsampling depth maps at low resolution, these methods can effectively expand the receptive field while decreasing the cost volumes. However, these methods typically require substantial memory and time consumption. To circumvent this, several methods like ~\cite{PatchMatchNet, Iter-MVS} attempt to employ lightweight gated recurrent units (GRU) to encode the depth probability distribution for each pixel. 
However, the adoptation of GRU  makes capacity for spatial recognition becomes constrained, thereby exhibit limitations in reconstructing textureless areas.
Other methods, like ~\cite{DELS-MVS, EPP-MVSNet} introduce an epipolar module to assemble high-resolution images into limited-size cost volumes. However, they often exhibit limited generalization, inhibiting their application in reconstructing scenarios with diverse characteristics.

\begin{figure}
\centering\includegraphics[width=0.95\textwidth]{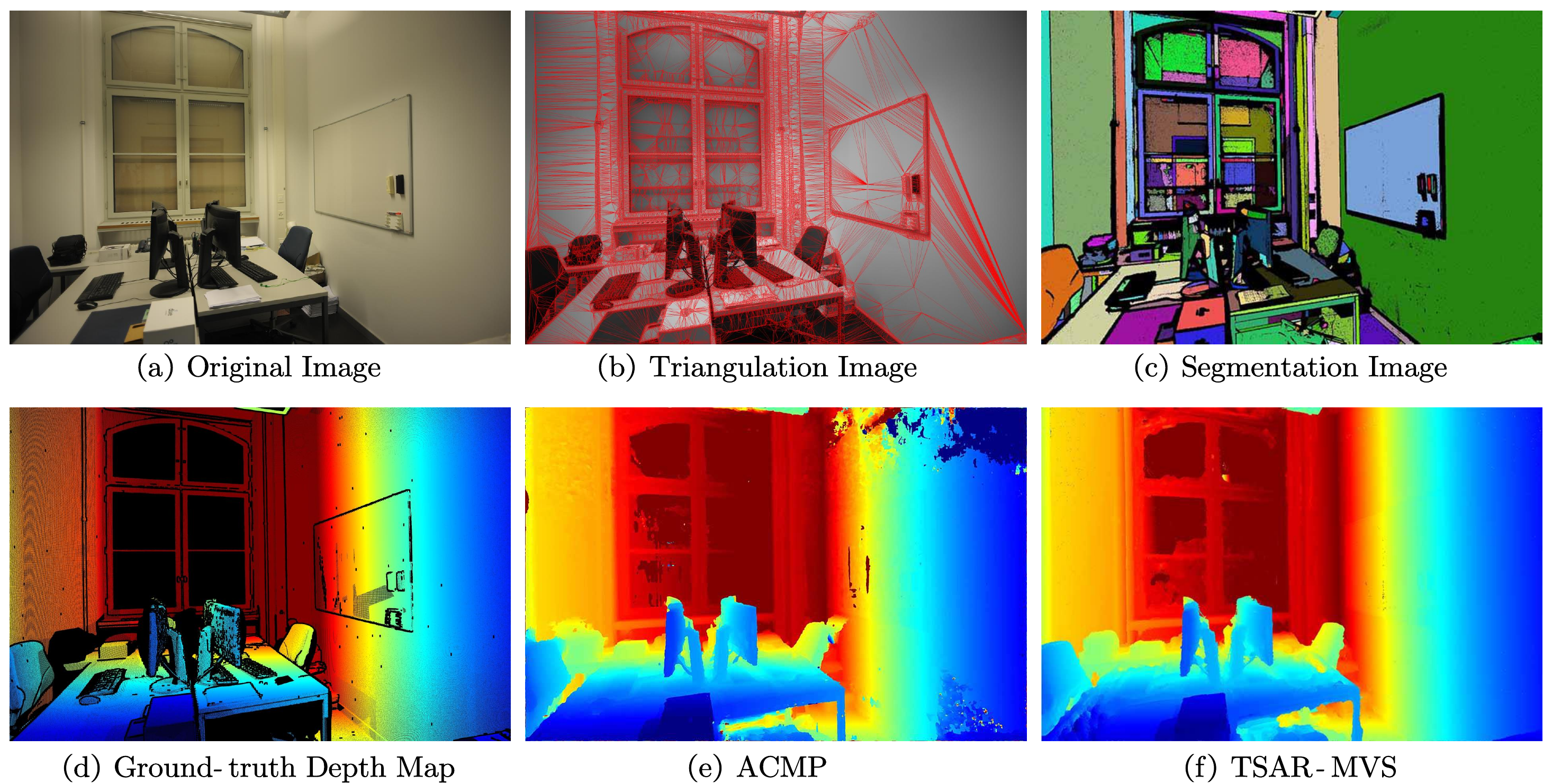}
\captionsetup{labelfont={color=black}}
\caption{\textcolor{black}{Comparative analysis between ACMP and our method. ACMP \cite{ACMP} (e) yields inferior reconstruction results owing to incomplete and noise-sensitive triangulation segmentation (b). 
Differently, textureless-aware segmentation (c) provides reliable guidance for TSAR-MVS (f) to planarize textureless areas.}
}
\label{fig1}
\end{figure}

\textcolor{black}{Large man-made scenarios are frequently characterized by textureless areas across varying scales, which can typically be predicted through their surrounding reliable pixels.
Therefore, an inspiring idea is to identify textureless areas across different scales and extract their reliable pixels for plane fitting, thereby achieving both reconstruction of textureless areas and refinement of detailed regions.}


\textcolor{black}{Therefore, we propose TSAR-MVS, a generalized post-processing MVS method that sequentially applies confidence-based outliers filtering, superpixel-based local refinement, and feature-based area generation to planarize textureless areas. Its distinctive design enables it to be applied after most methods and enhance their reconstruction results. 
As shown in Fig. \ref{fig1}, although ACMP~\cite{ACMP} constructs triangulation planar priors, it not only heavily relies on the precision of each reliable pixel but also struggles from incomplete image coverage. Differently, our method can exploit more robust and comprehensive plane prior insight, thus can recover precise and complete planes for reconstruction.
Moreover, unlike TAPA-MVS~\cite{TAPA-MVS} which are limited to planarizing localized areas, our textureless-aware segmentation can extract textureless areas across broader scales, thereby not limited by superpixels size. }

Specifically, we first introduce a joint hypothesis filter consisting of a confidence estimator and a disparity discontinuity detector to provisionally remove outliers. Inspired by~\cite{Li, DeepC-MVS}, we first combine three confidence features extracted from the matching cost to filter wrong estimates. Furthermore, we maintain areas with continuous disparity, underpinning the assumption brought by ~\cite{SGM}. Through integrating both confidence estimator and disparity discontinuities into an aggregation score, we can treat small disparity discontinuities as outliers, while preserving reliable object edges which are continuous in disparity.

Subsequently, to propagate the confident pixels with correct depth, we propose an iterative correlation refinement procedure. In each iteration, we first cluster pixels with similar perceptual features into same superpixels, then we apply RANSAC to planarize each superpixel, resulting in pixels within a superpixel sharing a consistent depth. Following this, a pixel-wise weighted median filtering (WMF) is used to propagate the correct depth to the surrounding areas. This iterative process is conducted over several rounds to ensure the propagation of reliable depth. Finally, another WMF is adopted to interpolate the depths and normals of the remaining pixels. 

Additionally, we propose a textureless-aware segmentation method combining edge detection and line detection.
Typically, in man-made scenarios ~\cite{ETH3D, strecha, TNT}, pixels within the same textureless area exhibit a high degree of similarity. 
Thus, we attempt to separate textureless areas and independently estimate the depth of their homogeneous surfaces. Specifically, we first apply the Roberts edge detection to extract high-frequency information, followed by the Hough line detection that enhances region discrimination by concatenating separated edges into lines. By aggregating residual pixels into miscellaneous areas, we can distinguish and further planarize them, thereby improving reconstructing result.

Experiments on extensive datasets, including the ETH3D high-resolution dataset, Tanks \& Temples dataset and Strecha dataset demonstrate that our method significantly outperforms other MVS methods. Furthermore, we demonstrate that our method can be incorporated as a component into other reconstruction pipelines to improve performance.

In summary, our contributions are as follows:

1)We propose a novel joint hypothesis filter by adaptly merging a confidence estimator and disparity discontinuities to yield an aggregation score that effectively eliminates outliers.

2)We present an iterative correlation refinement procedure that iteratively employs RANSAC and WMF to refine pixel estimates based on the surrounding confidence neighbors.

3)We introduce a textureless-aware segmentation technique that merges edge and line detection, thereby achieving the discrimination and planarization of textureless areas.

\textcolor{black}{4)Through extensive experiments, we verified that our method achieves state-of-the-art results and possesses strong generalization capability. }

\section{Related Work}
Conventional MVS methods can be categorized into four types, voxel based approaches~\cite{Volumetric, Ulusoy_2017_CVPR}, deformable polygonal meshes based approaches~\cite{Silhouette, Preserving}, patch based approaches~\cite{Locher_2016_CVPR, Collections} and depth maps based approaches~\cite{Schonberger_2016_CVPR, COLMAP}. Voxel based approaches enclose the scene with a bounding box and divide it into voxel grids. Polygonal meshes based approaches rely on a trusted initialization guess to iteratively improve accuracy. Patch based approaches require the textured surface to generate reliable 3D patches, which usually result in semi-dense reconstruction results for low-textured surface. Depth map based approaches progressively estimate corresponding depth maps and then perform the depth fusion for point cloud generation. 
Here we only review related depth maps based approaches and discuss methods concerned with PatchMatch Multi-View Stereo and Superpixel.

Deep-learning based MVS methods~\cite{MAR-MVS, EPP-MVSNet, DELS-MVS, Vis-MVSNet} that train learnable cost volumes and 3D regularization for reconstruction have also been proposed. 
Yao et al.~\cite{MVSNet, Rec-MVS} introduced a novel architecture featuring learnable cost volume through a 3D regularization strategy and further regularized cost along the depth direction via the GRU. 
Gu et al.~\cite{Cas-MVSNet} further proposed a cascaded refinement framework with a coarse-to-fine strategy to enhance robustness in reconstruction. 
Xu and Tao~\cite{PVSNet} designed a visibility network for prediction of neighboring images and an anti-noise training strategy for disturbing views. 
Furthermore, Xu et al.~\cite{SMU-MVSNet} exploits suppositional mesh for unsupervised MVS, significantly improving reconstruction through single-view occlusion reasoning and geometric consistency enforcement.
Moreover, Xu et al.~\cite{AGG-CVCNet} introduces a self-supervised MVS method that utilizes a geometry-guided cost volume completion technique to enhance depth map accuracy.
However, their growing complexity of network architectures requires a large amount of training data and can be prone to degradation on target domains that are distributed differently from the training set. 

\subsection{PatchMatch Multi-View Stereo}
Barnes et al.~\cite{PM} proposed the crucial theory of PatchMatch by adopting random initialization, propagation and refinement to search for approximate image patch pairs. 
Bleyer et al.~\cite{PMS} first proposed the PatchMatch Stereo that estimates 3D planes for each pixel in the disparity domain, and extended the spatial propagation to both view and time propagation. 
Galliani et al.~\cite{Gipuma} optimized sequential propagation scheme into a checkerboard diffusion propagation scheme. 
Sch{\"o}nberger et al.~\cite{COLMAP} further achieved concurrent depth refinement and 3D fusion by jointly predicting depths and normals.

Additionally, Xu and Tao~\cite{ACMM} developed multi-hypothesis view selection for cost aggregation and attempted to restore textureless areas through a multi-scale scheme. To address weak textures, Xu and Tao~\cite{ACMP, ACMMP} further introduced planar prior PatchMatch by introducing a probabilistic graphical model to optimize cost aggregation.
Moreover, Wang et al.~\cite{PatchMatchNet} integrated PatchMatch into a network and employed a learnable adaptive module for adaptive cost computation.
\textcolor{black}{Furthermore, Wang et al.~\cite{MG-MVS} proposed a mesh-guided approach within a pyramid architecture, enhancing reconstruction by leveraging coarse-scale surface mesh to refine finer-scale textureless areas.}
Yet limited by window size, the above methods are still not ideal in estimating textureless areas, nor can they adequately leverage relevance between neighbors to dislodge erroneous estimates.

\subsection{Superpixel}
Gouveia et al.~\cite{Confidence} first adopted a superpixel-based partition strategy to estimate depths and filtered reliable pixels with a random forest classifier. 
Furthermore, Zhang et al.~\cite{MESH} classified images into superpixels and then formulated a two-layer Markov Random Field for alignment.
Romanoni et al.~\cite{TAPA-MVS} employed SEED across two scales and introduced texture coefficients for different assignments.
Kuhn et al.~\cite{PCF-MVS} further achieved aggregation of textureless areas with a region-growing superpixel approach. 

Recently, Xue et al.\cite{XueOSGS19} employed a Markov Random Field (MRF) based on superpixels to compatibly simulate the correlation between edge depth and superpixel depth. 
Jung and Han\cite{JungH21} overcame the occlusion in 3D reconstruction by projecting superpixels onto adjacent viewpoints and performing extrapolation on plane models.
Additionally, Huang et al.~\cite{HuangHFZXLXC23} integrated the mean-shift clustering approach with superpixels to generate plane priors for textureless planar regions without using semantic information.
However, most approaches heavily depend on superpixel size or classification. Additionally, directly assigning pixels to their estimated superpixel planes may cause detail distortion.


\begin{figure*}
\centering
\includegraphics[width=\textwidth]{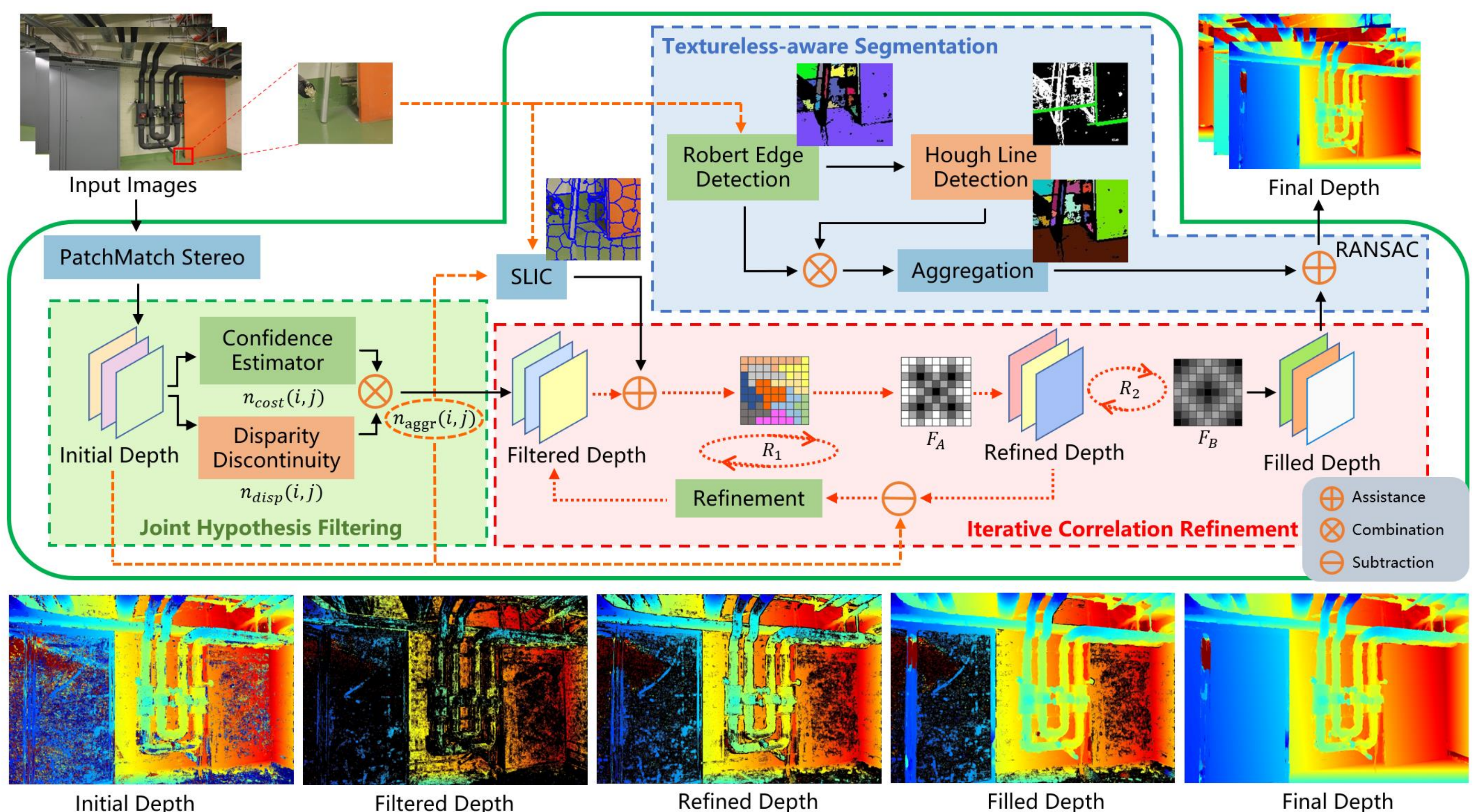}
\captionsetup{labelfont={color=black}}
\caption{An illustrated pipeline of our proposed method. We first acquire depth maps generated from Gipuma with adaptive checkerboard propagation. 
Subsequently, the joint hypothesis filtering merges a confidence estimator with a disparity discontinuity detector to eliminate incorrect depth estimations. 
Next, the iterative correlation refinement enhances the reliability of pixel estimates through two distinct iteration procedure.
The first iteration $R_{1}$ is conducted by repeatedly integrating RANSAC and WMF $F_A$ on a superpixel basis for pixel renewal.
Subsequent iteration $R_{2}$ is executed by adopting another WMF $F_B$ for interpolation of the remaining pixels.
Furthermore, a textureless-aware segmentation is utilized by integrating the Roberts edge detection and the Hough line detection for planarization of textureless areas.}
\label{fig2}
\end{figure*}

\section{Overview} 
\label{section:Overview}
Given all input images $\mathcal{I}=\{I_i \mid i=1 \cdots N\}$ and their corresponding camera parameters $\mathcal{P}=\{P_i \mid i=1 \cdots N\}$, our method amis to predicting their depth maps $\mathcal{D}=\{D_i \mid i=1 \cdots N\}$. 
The objective is to reconstruct the depth map for the reference image $I_{\mathrm{ref}}$, selected sequentially from $\mathcal{I}$, through pairwise matching with the remaining source images $I_{\mathrm{src}}(\mathcal{I} - I_{\mathrm{ref}})$.

An overview of our method is graphically represented in Fig. \ref{fig2}. We initiate the reconstruction procedure by using a baseline PatchMatch Stereo with adaptive checkerboard propagation~\cite{ACMM} to estimate initial depth maps for each image, which adopts the bilateral weighted adaptation of normalized cross-correlation~\cite{COLMAP} to compute the photometric cost. Subsequently, we deploy a joint hypothesis filter by fusing a confidence estimator with discontinuous disparity to construct the aggregated score, thereby eliminating erroneous estimates.

Subsequently,  we further apply an iterative correlation refinement to enhance the reliability of pixel estimates via a dual-phase iterative procedure.
The premier iteration is performed by repeatedly applying RANSAC-based planarization and WMF on a superpixel basis to retain more reliable pixels.
Then, for the remaining unreilable pixels, we continuously leverage another WMF to assign depths and normals to neighboring pixels in the subsequent iteration. 

Finally, we introduce a textureless-aware segmentation to planarize large textureless areas. 
We utilize the Roberts edge detection to extract high-frequency information, then connect the subtle edges with line segments using the Hough line detection. 
After grouping the remaining pixels into miscellaneous regions, we achieve the identification of textureless areas and then carry out RANSAC-based planarization. 

\begin{figure}
\centering\includegraphics[width=0.95\linewidth]{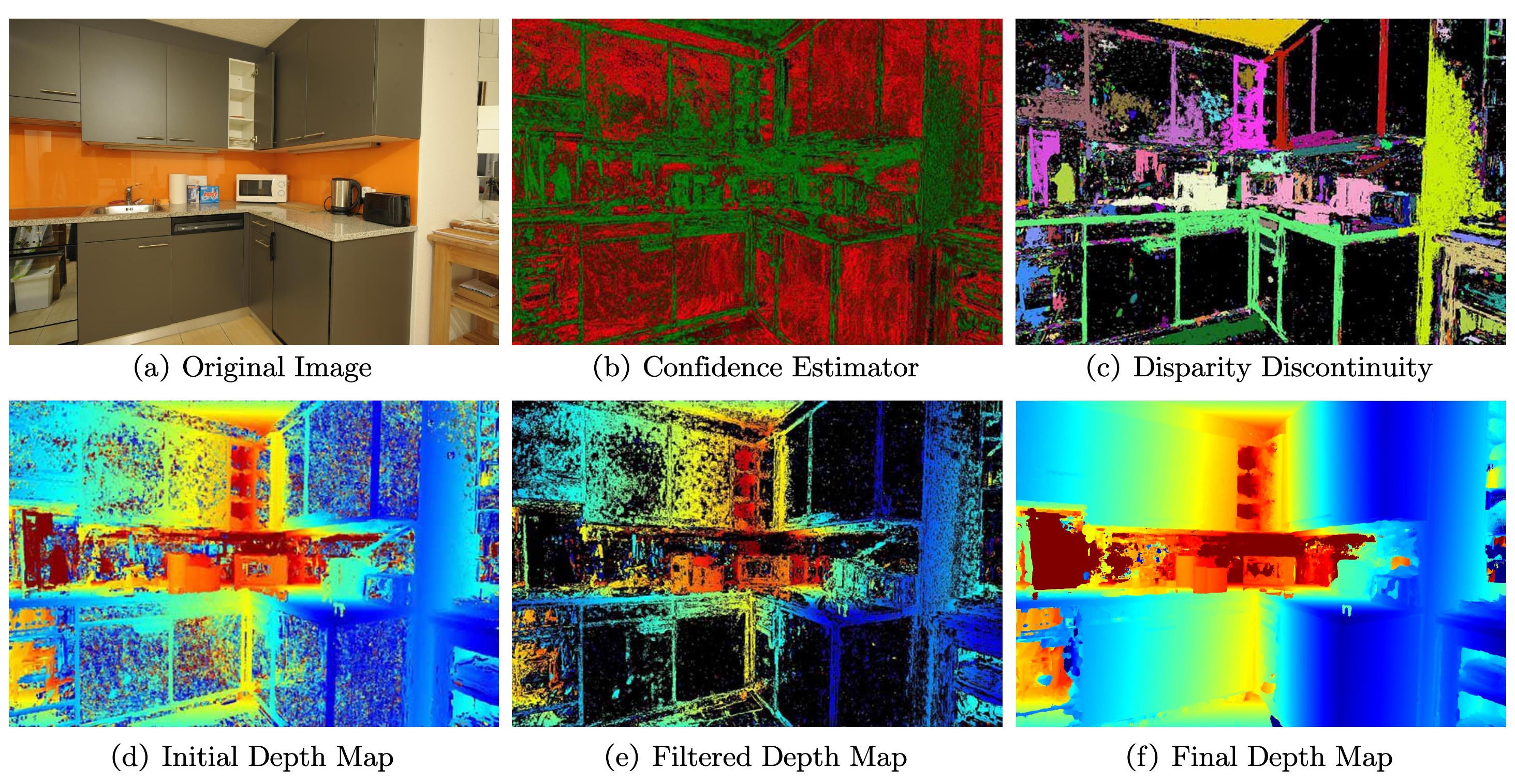}
\captionsetup{labelfont={color=black}}
\caption{\textcolor{black}{Joint hypothesis filtering.} Pixels satisfying $n_{cost}(i,j) > 1$ in (b) are denoted in green, whereas others are denoted in red for visualization. Pixels associated with adjacent disparities in (c) are connected and visualized using the same color. 
Comparing the initial depth map (d), the filtered depth map (e) that merges confidence estimator (b) and disparity discontinuity (c) retains a notable portion of reliable depths.}
\label{fig3}
\end{figure}

\section{Method} 
\subsection{Joint Hypothesis Filtering}\label{section:Joint Hypothesis Filter}
In this subsection, we present the joint hypothesis filter, a key component of our method designed to eliminate outliers and discontinuous disparity peaks, thereby preparing the data for subsequent processing steps. As illustrated in Fig. \ref{fig3}, our method integrates a confidence estimator and a disparity discontinuity filter to calculate the aggregated score, thereby effectively removing unreliable pixels. Distinct from previous methods~\cite{TAPA-MVS, PCF-MVS} that permanently discard pixels, our pipeline allows for the potential restoration of discarded pixels, thus preserving more possibilities based on the practical context of subsequent refinement.

\subsubsection{Confidence Estimator}
Inspired by~\cite{Confidence}, we determine the confidence based on three distinct features related to the matching cost.
Specifically, the total confidence $n_{cost}(i,j)$ is composed of three different confidence measures. Given a pixel $(i,j)$ in the reference image $I_{ref}$ with its corresponding pixel $(i',j')$ in the source image $I_{src}$, we synthesize the total confidence $n_{cost}(i,j)$ as follows:
\begin{equation}
\begin{split}
n_{cost}(i,j)=2-\frac{1}{2}c_{min}(i,j)+\frac{c_{min}(i,j)}{c_{sec}(i,j)}
\\
-\left| c_{min}^{ref}(i,j)-c_{min}^{src}(i',j') \right|
\end{split}
\end{equation}

The first part $c_{min}(i,j)$ represents the minimal matching cost, since a lower cost generally indicates a higher degree of relevance and thus a more accurate depth prediction. The second part $c_{min}(i,j)/c_{sec}(i,j)$ is calculated by the ratio between the minimal and sub-minimal cost. This feature is introduced to detect errors induced by ambiguity during matching.
The last part $|c_{min}^{ref}(i,j) - c_{min}^{src}(i',j')|$ denotes the absolute difference between the pixel with the optimal cost in the reference image and its corresponding pixel in the source image, which is often referred to as Left-Right Difference (LRD).

\subsubsection{Disparity Discontinuity}
\label{section:Disparity Discontinuity}
Isolated outliers frequently present themselves as minor peaks, with depths markedly differing from surrounding pixels. 
However, utilizing a fixed depth threshold for filtering often proves inadequate in distinguishing scenes with varying depth ranges.
Consequently, we employ disparity filtering instead of depth filtering. Moreover, inspired by ~\cite{SGM}, we upgrade the original 4-connected grid into an 8-connected grid to reduce mislabeling. 
Additionally, our strategy merges multi-scale disparity maps and dynamic thresholds for superior results.

Specifically, each disparity map is downsampled into 3 layers. At the $k^{th}$ level, we first formulate connected domain by connecting pixels whose disparity variance with neighbors is less than $t_d$. Subsequently, we preserve connected domains whose number of pixels surpass $t_n$. Therefore, the final disparity discontinuity confidence for each pixel is calculated by: $n_{disp}(i,j) = \sum_k{n_k / k}$, wherein $n_k$ equals 1 if the pixel is preserved at the $k^{th}$ level, and 0 otherwise. In experiment, we respectively set $t_d$ and $t_n$ as $3 \times 2^k$ and $5 \times 2^k$.

\subsubsection{Aggregated Score}
\label{section:Aggregated Score}
By fusing the aforementioned cost-driven confidence estimator and the disparity discontinuity detector, the aggregated score $n_{aggr}(i,j)$ is defined by:
\begin{equation}
\begin{split}
n_{aggr}(i,j) = \tau \times n_{cost}(i,j) + (1 - \tau) \times n_{disp}(i,j)
\end{split}
\end{equation}
This unified score holistically evaluates credibility, incorporating both individual pixel costs and inter-pixel disparity correlations. 
Pixels $(i,j)$ that satisfy $n_{aggr}(i,j) > \eta$ are characterized as confident pixels with reliable initial depth estimates. 
Note that $n_{aggr}(i,j)$ is also leveraged in subsequent superpixel segmentation and pixel refinement.

\begin{figure*}
\centering
\includegraphics[width=\linewidth]{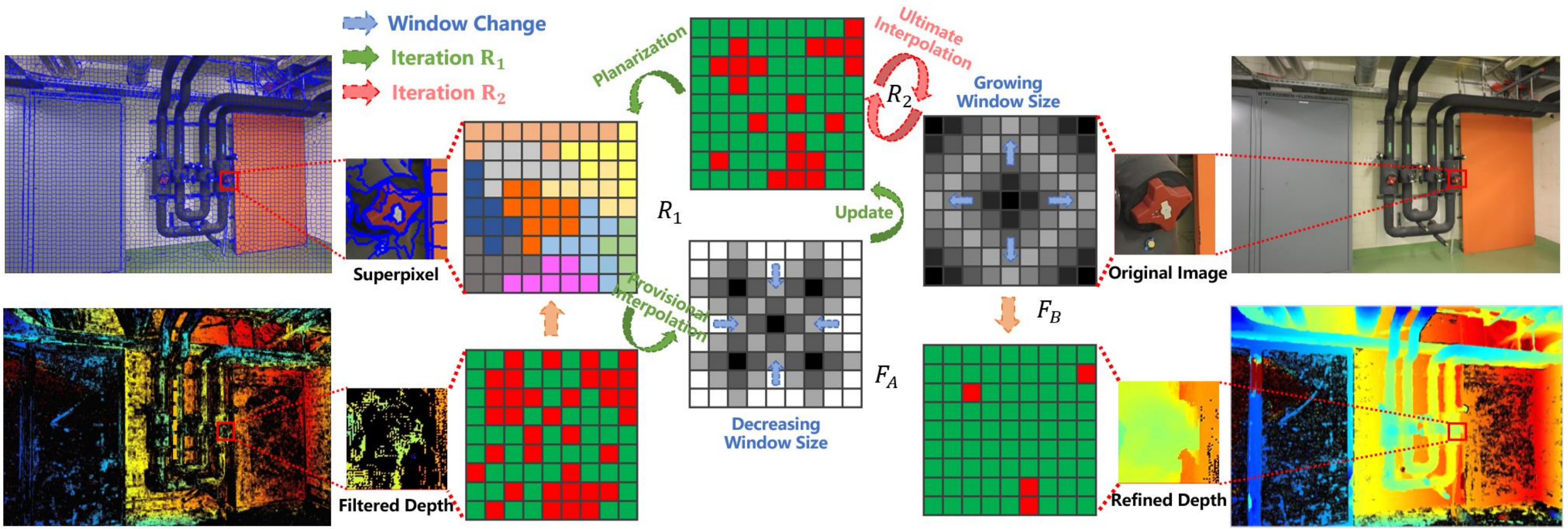}
\caption{Illustration of the proposed iterative correlation refinement. 
The framework is fundamentally structured around two discrete iterative stages, $R_1$ and $R_2$.
Specifically, after conducting joint hypothesis filtering, reliable pixels with green color are selected as the initial input for superpixel planarization within the first stage $R_1$.  
Throughout each iteration, the pixels are overlaid by the depth of planes predicted by their superpixels, and subsequently interpolated using $F_A$ with the window size gradually decreasing according to the number of iterations. We then refine the pixels by contrasting them with the initial depth map, readying them for the next round of input. 
Progressing to the subsequent stage $R_2$, we employ $F_B$ with its window size incrementally expanding to interpolate the remaining pixels. 
}
\label{fig4}
\end{figure*}

\subsection{Iterative Correlation Refinement}
Most superpixel-based methods~\cite{MESH} commonly use RANSAC to planarize the pixels in the generated superpixels, considering them to have the same depth estimates. 
This approach has two potential drawbacks. 
First, simply replacing the depth estimates of the pixels with those of the superpixels can cause distortion, as superpixels tend to represent holistic views, while individual pixels contain specific details. 
Second, RANSAC has stringent requirements on data distribution, meaning that the filtering may cause inaccuracies in depth once the data distribution is beyond satisfaction.

To address these issues, we introduce an iterative correlation refinement procedure as illustrated in Fig. \ref{fig4}, which is structured around two divergent iterative phases. 
In the preliminary phase, given a preliminary disparity map calculated by PMS mentioned in \ref{section:Overview}, we first use reliable pixels as input for planar prediction, assigning pixels with the correctly-predicted depths of their superpixels.
Subsequently, we employ weighted median filtering (WMF) to propagate the correct depth to neighboring pixels. 
Following multiple iterations of this mechanism, we evaluate the disparity between the refined disparity $dp^{est}$ and the initial disparity $dp^{org}$.
Subsequently, in the following phase, we repeatedly utilize another WMF for the interpolation of the remaining unreliable pixels.

\subsubsection{Superpixel}

The Simple Linear Iterative Clustering (SLIC) algorithm ~\cite{SLIC_sota} is employed for image segmentation in our method, which is performed using k-means pixel clustering. However, our innovation lies in combining this with our aggregated score. Specifically, the distance between the pixel $p$ and the superpixel $c_k$ is given by:
\begin{equation}
\begin{split}
D^{\prime}=\sqrt{\left(\frac{d_c}{10}\right)^2+\left(\frac{d_s}{20}\right)^2 + n_{avr}}
\end{split}
\end{equation}
where $d_c$ and $d_s$ respectively depict the color and spatial distance between $p$ and $c_k$, and $n_{avr}$ characterizes the mean aggregation score $n_{aggr}$ of all pixels inherent to each superpixel $c_k$. 
Such an integration inclines towards increasing superpixel size within textureless areas where reliable pixel distributes more sparsely, thereby facilitating their planarization.

The number of superpixel is also pivotal in achieving precise imagery. 
Specifically, insufficient superpixels could lead to excessive pixels being planarized together, thereby resulting in inaccuracy in depth details and loss of valuable information.
Conversely, too many superpixels can pose challenges for reconstruction, as the data may be insufficient for the RANSAC algorithm to function effectively. 
\textcolor{black}{Empirically, we opt to set the number of superpixel such that the size of each superpixel $S_{sp}$ is approximately 400 in our images.}

\subsubsection{Random Sample Consensus}
\textcolor{black}{RANSAC iteratively estimates the overall distribution from a set of data with outliers. In this work, we achieve planarization by conducting RANSAC on reliable pixels within each superpixel.
Note that the distance threshold and the credibility threshold play pivotal roles during RANSAC. The former determines whether each pixel can be considered an inlier of the estimated plane, while the latter uses the number of inlier to judge the reliability of estimated plane.
To improve prediction accuracy, we propose an dynamic distance threshold $t_k$ and introduce an adaptive ratio $r_{k}$ to replace the previous credibility threshold.}

\textcolor{black}{Specifically, we first seperate image into $N_{sp}$ superpixels, thus we have all superpixels $\mathcal{P}=\{P_i \mid i=1 \cdots N_{sp}\}$. 
Then the adaptive ratio $r_{k}$ that evaluates reliability is defined as the quotient of the number of reliable pixels and the superpixel magnitude. }
However, the reliability of the estimated plane can be influenced by other factors such as data distribution. 
Therefore, solely using $r_k$ for depth estimation would be imprecise.
To address this, we also require the depth of the predicted 3D plane to be within an adaptive range defined by the minimum and maximum depth estimates of the pixels in the superpixel. 
Formally, we denote a superpixel set that satisfies both parts as $S$, defined by:
\begin{equation}
\label{frc}
\begin{array}{c}
	S\in \{P_k|r_{k}>\sigma \land D_k\in \left[1.2 \cdot D_{min},0.8 \cdot D_{max} \right]\}
\end{array}
\end{equation}

where $k \in \{1,2,...,N_{sp}\}$, $D_k$ denotes the predicted depth at the centroidal position of the $k$-th superpixel, $D_{min}$ and $D_{max}$ respectively represent the minimum and maximum depth estimates of the pixels in the superpixel.
\textcolor{black}{Furthermore, for each superpixel $P_k$, the dynamic distance threshold $t_k$ to discern if its internal pixels can be classified as inliers is defined as:}
\begin{equation}
\label{adt}
t_k=\begin{cases}
	D_k \cdot 10^{-3} & , \mathrm{\text { if } } P_k \in S\\
	5 \cdot 10^{-3} & ,\mathrm{else}\\
\end{cases}
\end{equation}

\subsubsection{Weighted Median Filtering}
Our method incorporates two similar weighted median filters, each with distinct objectives. 
The first filter is utilized in the first iteration $R_1$ to provisionally retain pixels, while the second is adopted in the subsequent iteration $R_2$ to ultimately interpolate the remaining irrational pixels. 
Both filters calculate weights based on spatial distance and color variance.
\textcolor{black}{Specifically, for any pixel $p$, the weight $w$ is defined as follows:}
\begin{equation}
\label{weight}
w=\sum_{q \in W_p}e^{ -\left[ \frac{\sqrt{(L_p-L_q)^2}}{\alpha^2}+\frac{\left|C_p-C_q\right|}{\beta^2} \right] }
\end{equation}
where $q$ denotes the pixels within the window $W_p$ centered on $p$, $\sqrt{(L_p-L_q)^2}$ represents the Euclidean distance between the positions of pixels $p$ and $q$, $\left|C_p-C_q\right|$ represents the color difference between pixels $p$ and $q$ in the RGB space, and $\alpha$ and $\beta$ are the corresponding spatial and color coefficients.
\textcolor{black}{Additionally, to provide the window with a more suitable receptive field, we propose adaptive window sizes $l_A$ and $l_B$ for both filters $F_A$ and $F_B$, defined as:}
\begin{equation}
\label{window}
l_A= \delta \cdot 2^{(N_t - N_c)}, l_B= \delta \cdot 2^{N_c}
\end{equation}
\textcolor{black}{where $N_t$ and $N_c$ respectively represent the total number of iterations and the current number of iterations.} Therefore, the window size of $F_A$ decreases exponentially as iterations progress, thereby increasing its focus on the local perception field. 
In contrast, the window size of $F_B$ grows exponentially with each iteration, enabling the propagation of authentic pixels to a greater extent.
Hence, in comparison to TAPA-MVS~\cite{TAPA-MVS}, which employs the bilateral weighted median of neighboring pixels, our filters exhibit a more problem-specific focus for refinement and afford greater flexibility for interpolation.

\textcolor{black}{However, adaptive window sizes can lead to an exponential increase in runtime. To ensure that the runtime remains consistent across iterations, we set the interval sizes of both filter to be $1/\delta$ times of their window sizes, such that the time complexity of the filtering is maintained at $O(\delta^2)$ and remains independent of the number of iterations.}

\subsubsection{Iteration procedure}
The proposed refinement involves two distinct iterations, $R_1$ and $R_2$.
In the first iteration $R_1$, we employ RANSAC and $F_A$ to propagate reliable depth values. 
RANSAC is initially applied to planarize each superpixel, which allows each pixel $(x, y)$ to be assigned its correctly-predicted superpixel depths $D_{sp}$. 
The filter $F_A$ is subsequently utilized to further propagate these reliable depths, leveraging the local consistency of superpixels while also preserving the specific details of individual pixels. 
Specifically, given the discussions in Section \ref{section:Disparity Discontinuity}, where we found the disparity threshold to be more suitable for evaluation than the depth threshold, we convert the propagated depth $d$ to a disparity $dp$ using the following equation:
\begin{equation}
\label{dp}
dp =-\frac{1}{\vec{n}_z}\left(d+\vec{n}_x x+\vec{n}_y y\right)
\end{equation}
where $d=-\vec{n}^{\mathrm{T}} \left[x, y, dp\right]^{\mathrm{T}}$ denotes the distance to the camera origin, $\vec{n}$ represents the normal vector.
After converting depth to disparity, we retain pixels satisfying following condition:
\begin{equation}
\label{ths}
\left|dp^{est} - dp^{org} \right| < \frac{\mu}{n_{aggr} \times N_c}
\end{equation}
where $dp^{est}$ and $dp^{org}$ respectively denote the refined disparity gained from the first filter $F_A$ and the initial disparity gain from the modified Gipuma. 
Meanwhile, $n_{aggr}$ represents the aggregated cost gained from \ref{section:Aggregated Score}.
Pixels whose depth values satisfy the Eq. \ref{ths} are deemed trustworthy, and thus, serve as the new input for the ensuing iteration within $R_1$, while others are provisionally discarded. 
Successive refinement rounds gradually enhance the reliability of depth predictions, both for individual pixels and their superpixels.

Upon the initial iteration reaches termination, the second iteration $R_2$ is subsequently performed. 
This procedure consistently employs another WMF, $F_B$, to interpolate the depth of the residual irrational pixels by referencing the depth values of adjacent reliable pixels.  
Ultimately, through this dual-phase iterative process, we can achieve the refinement of erroneous estimates via their adjacent trustworthy pixels.

\subsection{Textureless-aware Segmentation}
After the iterative correlation refinement, most superpixels within well-textured area can achieve depth refinement. However, due to the absence of reliable pixels in large textureless areas, iterative correlation refinement cannot leverage neighboring pixels for optimization. Therefore, a large textureless areas detection method is urgently needed.

In man-made scenarios, textureless areas like walls and floors are typically enclosed by straight lines, while well-textured areas contain feature points. Then an inspiring idea is to first use edge detection to identify feature points, and then connect these points into lines. Therefore, areas enclosed by these lines may contain textureless areas with varying sizes.

\begin{figure*}
\centering
\includegraphics[width=\textwidth]{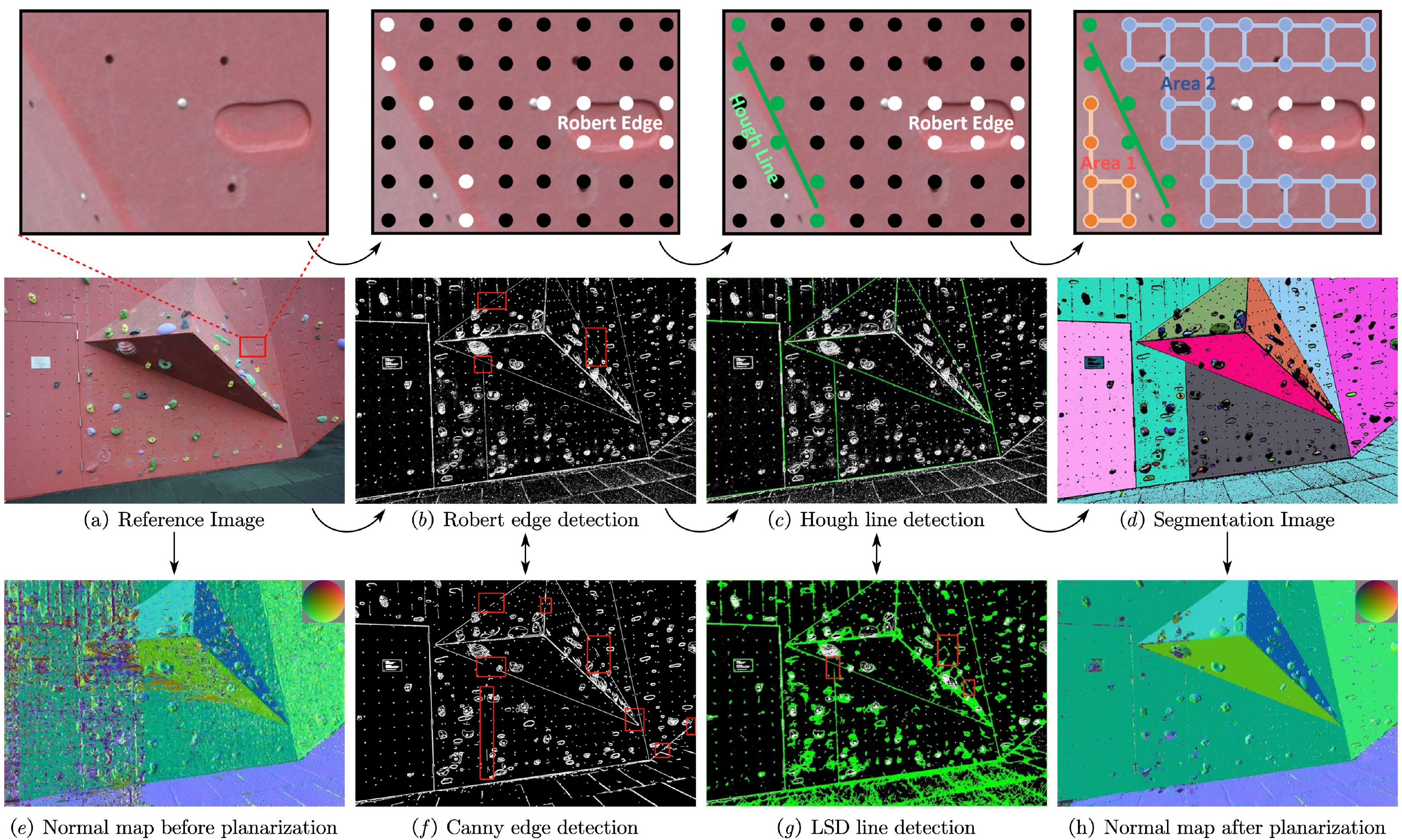}
\captionsetup{labelfont={color=black}}
\caption{\textcolor{black}{Textureless-aware segmentation.} Note that red boxes in (b), (f) and (g) contain edges that remain undetected and green segments present in (c) and (f) represent detected lines. 
Additionally, diverse color palettes within (d) signify distinctively detected planes.
Comparing (b) and (f), it is apparent that the Roberts detection outperforms the Canny detection in terms of extracting more edges, thereby facilitating plane segmentation.
Comparing (c) and (g), it is evident that the Hough detection can successfully connect fractured edges which are missing within red boxes in (b), while the LSD fails to distinguish these necessary edges and introduces redundant connections.
Consequently, we opt for the Roberts detection and the Hough detection to yield our composite segmentation image in (d). 
Comparing (e) and (h), we find that the textureless-aware segmentation can not only complete textureless areas but also preserve fine detail.}
\label{fig5}
\end{figure*}

Therefore, we propose a textureless-aware segmentation method across superpixel segmentation, as depicted in Fig. \ref{fig5}. 
Specifically, we first utilize the Roberts edge detection to extract edges from reference images. 
Then we employ the Hough line detection to connect discontinuous Robert edges into lines, thereby enhancing the distinction between areas.
Subsequently, we utilize the two-pass labeling algorithm ~\cite{two-pass} to merge the remaining undetected pixels and lines into distinct regions.
Finally, we perform planarization according to area size and further verify its correctness to reconstruct large textureless areas. 

\begin{figure*}
\centering
\includegraphics[width=\linewidth]{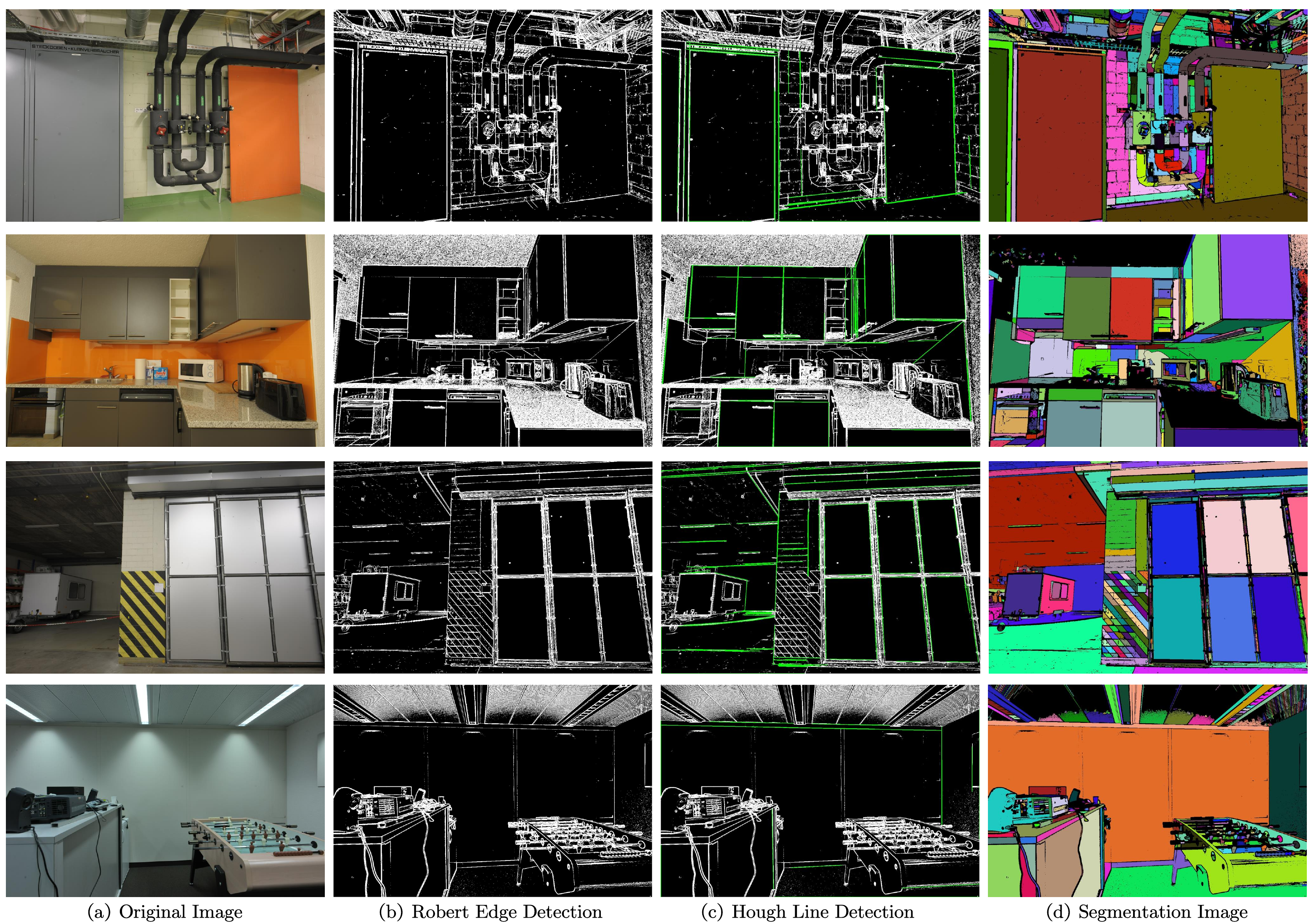}
\caption{\textcolor{black}{More visualization results of the textureless-aware segmentation on partial scenes of ETH3D datasets (\emph{pipes}, \emph{living room}, \emph{delivery area} and \emph{kicker}). Detected edges are highlighted in white in (b) and (c), detected lines are highlighted in green in (c). Different color in (d) denoted different connected areas.
}
}
\label{SE}
\end{figure*}

\textcolor{black}{
More visualization results in representative scenes is shown in Fig. \ref{SE}. As can be seen, the combination of the Robert edge detection and the Hough line detection can effectively extract texturelesss areas across varying scales as intended. Moreover, despite the presence of lighting variations within textureless areas of pipes and kicker scenes, our method can precisely extract them as intended, demonstrating its robustness against illumination disturbances.
}

\subsubsection{Roberts Edge Detection}
To handle subtle changes in color gradient, an edge detection algorithm with high sensitivity to gradient variations is required, thereby we choose Roberts operator.
Compared with Canny, Sobel and Laplacian operators, Roberts operator is recognized for its ability to capture minor texture changes and slight color gradient variations.
Specifically, for each pixel $(x, y)$ in the reference images $I_{ref}$, the gradient $g(x, y)$ of grayscale $f(x, y)$ is defined by:
\begin{equation}
\begin{array}{c}
	g(x,y)=\left( \sqrt{f(x,y)}-\sqrt{f(x+1,y+1)} \right) ^2\\
	+\left( \sqrt{f(x+1,y)}-\sqrt{f(x,y+1)} \right) ^2\\
\end{array}
\end{equation}
Moreover, for image $I_i$ we have all pixels $\Omega_i \subset$ $\mathbb{R}^2 \rightarrow \mathbb{R}^{3}$.
Then we define the point set $E_{Robert}$ obtained through Robert edge detection as:
\begin{equation}
E_{\mathrm{Robert}}=\{(x, y)\in \Omega_i\mid g\left(x,y\right)>4\}
\end{equation}

\subsubsection{Hough Line Detection}
\textcolor{black}{After extracting edge through the Robert operator, a line detection algorithm that can effectively separate different textureless areas is required, thereby we choose Hough.}
As shown in Fig. \ref{fig5}, while other methods like Line Segment Detector (LSD) primarily focus on redundant short-edge connections, Hough line detection can effectively extract non-redundant edges across varying scales, thus more suitable for our analysis. 

\textcolor{black}{Specifically, we first utilize Hough transform to map all Robert edges in Cartesian coordinate system into lines in Hough space. 
Then we then identify intersection points that satisfy $n_l > \varepsilon$ in Hough space, where $n_l$ denotes the number of intersecting lines for each intersection point.
Therefore, intersection points that can be intersected by at least $\varepsilon$ lines in Hough space are exactly our desired Hough lines in Cartesian coordinate system.}

\textcolor{black}{Moreover, we observe that longer lines should tolerate larger blank interval between edges, enabling us to extract textureless areas across varying scales.
Therefore, we introduce an maximum connection threshold $t_c$ to connect edges within a certain distance, thereby avoiding the unintended connection of distant edges which may cause the fragmentation of textureless areas. Specifically, $t_c = \min \left\{ 0.1 \cdot l_s, \upsilon \right\}$, where $l_s$ denotes the length of line and $\upsilon$ is the truncation threshold to prevent an excessively large blank interval. 
Let $l_g$ represent the blank interval between two edges collinear with the Hough line to be extended, then the Hough line can only be lengthened when the blank interval $l_g$ between two edges is no less than $l_s$. Then the point set $E_{Hough}$ obtained through Hough line detection is defined as:
\begin{equation}
E_{Hough}=\left\{(x, y) \in \Omega_i \mid(x, y) \text { on lines with } n_l>\varepsilon \text { and } l_g \geq t_c\right\}
\end{equation}
}
\subsubsection{Pixels to Connected Regions}
\textcolor{black}{
After obtaining $E_{Robert}$ and $E_{Hough}$, we can combine them to form $E_{Total}$, defined as: }
\begin{equation}
E_{total} = E_{Robert} \cup E_{Hough}
\end{equation}
\textcolor{black}{Then, for any point $p$ and its neighbor $q$, if both $p$ and $q$ do not belong to $E_{Total}$, a connection between $p$ and $q$ is established:}
\begin{equation}
\text{connect}(p, q) = 
\begin{cases} 
1 & \text{if } q \notin E_{\text{total}} \land p \notin E_{\text{total}},\\
0 & \text{otherwise}.
\end{cases}
\end{equation}
\textcolor{black}{
Here, 1 indicates the connection between p and q, while 0 indicates no connection.
Ultimately, by connecting all satisfactory pixels, we can obtain numerous connected areas across various scales, which may include textureless areas we aim to identify.
To actualize this, we utilize a two-pass connected component labelling algorithm ~\cite{two-pass}, which aggregates all connected pixels into a region and assigns a unique label to each region. 
}

\subsubsection{Connected Regions to Textureless Areas}
\textcolor{black}{In this section, we will describe how to select our desired textureless areas for planarization among all connected regions generated before.}
Based on our previous discussion, we have divided the image into several connected regions, denoted as $\mathcal{R}=\{R_i \mid i=1, \cdots, M\}$. 
Each of these regions is considered to be an individual plane.
Since iterative correlation refinement is capable of addressing textureless areas at the superpixel level, our focus shifts to detecting larger textureless areas that refinement fails to address.
Therefore, we regard regions satisfying the following condition as textureless areas:
\begin{equation}
\label{unt}
R_{t}\in \left\{ R_j|N_j> \kappa \cdot 10^{4}, j=1,2,...,M \right\} 
\end{equation}

In this equation, $\mathcal{N}=\{N_i \mid i=1 \cdots M\}$ symbolizes the count of pixels within each region of the image. 
\textcolor{black}{
Moreover, we empirically choose $\kappa = 0.8$ because such threshold can effectively cover most of the large textureless areas that cannot be refined by the iterative correlation refinement, while also preventing unnecessary planarization on already refined small textureless areas, thereby avoiding increased time complexity.
Therefore, regions that encompass a sufficient number of pixels can be selected.
Subsequently, we employ the RANSAC algorithm for planarization and the least square algorithm for refinement. 
}

In order to prevent the erroneous planarization of textureless areas, we introduce an aggregate measure $\bar{D}$ to discern the connectivity between the predicted plane and its surroundings reliable pixels. Specifically, we initially extract all boundary points within each textureless area, termed as $P_{edge}$. Then for each pixel $p$ within $P_{edge}$, we compute the depth difference between itself $d_{p}$ and the average depth of all its adjacent reliable pixels $\bar{d_n}$.
Subsequently, we can determine $\bar{D}$ by calculating the mean depth difference:
\begin{equation}
\bar{D}=\frac{1}{N_{edge}}\sum_{p\in P_{edge}}{\left| d_p - \bar{d_n} \right|}
\end{equation}
where $N_{edge}$ is the total number of boundary points within each textureless area.
We deem the planarization of the textureless area as correct if $\bar{D}$ is less than $\gamma$; otherwise, misidentified planes will not be updated by its planarization result.

\section{Fusion}
We employ a fusion step to obtain the point cloud result, following the conventional PatchMatch pipeline from \cite{Gipuma, COLMAP}.
Specifically, we successively treat each image as a reference and convert its depth map into 3D points in world coordinates.
These points are subsequently projected onto neighboring images to generate corresponding matches. 
We consider a match to be consistent if it satisfies the following criteria: a relative depth difference less than $0.01$, an angle difference between normals less than $30°$, and a reprojection error of $\Psi \le 2$, as in \cite{COLMAP}. If the number of consistent matches across all neighboring views exceeds $1$, the average depth estimate is accepted. 
Ultimately, the corresponding 3D points and normal estimates of these consistent depth estimations are aggregated and averaged into a unified 3D point.

\section{Experiment}
Our method is evaluated on three prominent MVS datasets, the ETH3D high-resolution multi-view stereo benchmark ~\cite{ETH3D}, \textcolor{black}{the Tanks \& Temples dataset (TNT)} and the Strecha dataset~\cite{strecha}. 
The above-mentioned datasets are all categorized into datasets with unstructured viewpoints.
We initially conducted a qualitative and quantitative evaluation of our point cloud results against various baselines. 
Moreover, we conducted separate ablation studies for each proposed component and key hyperparameters to demonstrate the effectiveness of each module and the rationale behind our parameter selections.
The experimental results demonstrate that our approach not only significantly outperforms both traditional and learning-based methods but also exhibits strong generalization capability. 

\subsection{Dataset and Settings} \label{section:Dataset and Settings}
The ETH3D high-resolution multi-view stereo benchmark \cite{ETH3D} comprises 25 scenarios, each includeing images of $6, 221\times 4, 146$ resolution.
Its varying discrete viewpoints and wide variety of scene types pose a greater challenge for reconstruction. 
This dataset is divided into training and testing dataset. The training dataset with 13 scenarios publishes both ground truth point clouds and ground truth depth maps, while the ground truth for the testing set with 12 scenarios is retained by its benchmark site.

\textcolor{black}{The Tanks and Temples (TNT) benchmark encompasses 14 distinct scenes with the resolution of $1, 920\times 1, 080$ , which include individual objects like \textit{trains} and large indoor scenes like \textit{museums}. 
The ground truth depth maps were captured using high-quality industrial laser scanners.
Both the ETH3D and TNT benchmark are divided into training and testing dataset. 
We have uploaded our results to their official websites for public reference.}

The Strecha dataset \cite{strecha} contains six outdoor scenes. Among these, the Fountain and HerzJesu scenes provide ground truth point clouds, each comprised of 11 and 8 images respectively, with a resolution of $3, 072 \times 2, 048$. 

\begin{table}
  \centering
  \renewcommand{\arraystretch}{1.05} 
    \resizebox{0.8\linewidth}{!}{
        \begin{tabular}{c|ccc|ccc|ccc|ccc}
        \hline
        \multirow{3}{*}{Method} & \multicolumn{6}{c|}{Train} & \multicolumn{6}{c}{Test} \\
        \cline{2-13} & \multicolumn{3}{c|}{$2cm$} & \multicolumn{3}{c|}{$10cm$} & \multicolumn{3}{c|}{$2cm$} & \multicolumn{3}{c}{$10cm$}  \\
        \cline{2-13} & Acc. & Comp. & F$_1$ & Acc. & Comp. & F$_1$ & Acc. & Comp. & F$_1$ & Acc. & Comp. & F$_1$ \\   
        \hline 
        \multirow{1}{*}{COLMAP\cite{COLMAP}} & \textbf{91.85} & 55.13 & 67.66 & \textbf{98.75} & 79.47 & 87.61 & \textbf{91.97} & 62.98 & 73.01 & 98.25 & 84.54 & 90.40 \\  
        \multirow{1}{*}{OpenMVS\cite{openmvs}} & 78.44 & 74.92 & 76.15 & 95.75 & 89.84 & 92.51 & 81.98 & 78.54 & 79.77 & 95.48 & 90.75 & 92.86 \\ 
        \multirow{1}{*}{IterMVS-LS\cite{Iter-MVS}} & 79.79 & 66.08 & 71.69 & 94.48 & 94.48 & 85.25 & 84.73 & 76.49 & 80.06 & 95.42 & 85.81 & 90.15 \\  
        \multirow{1}{*}{ACMP\cite{ACMP}} & 90.12 & 72.15 & 79.79 & 97.97 & 87.15 & 92.03 & 90.54 & 75.58 & 81.51 & \textcolor{red}{97.47} & 88.71 & 92.62 \\  
        \multirow{1}{*}{ACMMP\cite{ACMMP}} & \textcolor{red}{90.63} & 77.61 & 83.42 & 97.99 & 93.32 & 95.54 & \textcolor{red}{91.91} & 81.49 & 85.89 & \textbf{98.05} & 94.67 & 96.27	\\ 
        \multirow{1}{*}{APD-MVS\cite{APD-MVS}} & 89.14 & \textbf{84.83} & \textcolor{red}{84.83} & 97.47 & \textbf{96.79} & \textcolor{red}{97.12} & 89.54 & 85.93 & \textcolor{red}{87.44} & 97.00 & 96.95 & \textcolor{red}{96.95} \\
        \hline 
        \multirow{1}{*}{TAPA-MVS\cite{TAPA-MVS}} & 85.88 & 71.45 & 77.69 & 96.79 & 90.98 & 93.69 & 85.71 & 74.94 & 79.15 & 94.93 & 90.35 & 92.30\\  
        \multirow{1}{*}{PCF-MVS\cite{PCF-MVS}} & 84.11 & 75.73 & 79.42 & 95.98 & 90.42 & 92.98 & 82.15 & 79.29 & 80.38 & 92.12 & 91.26 & 91.56 \\  
        \multirow{1}{*}{MG-MVS\cite{MG-MVS}} & --- & --- & --- & --- & --- & --- & 80.32 & 87.11 & 83.41 & 94.08 & \textcolor{red}{97.24} & 95.61 \\ 
        \hline 
        \multirow{1}{*}{TSAR-MVS + GP.} & 85.32 & 82.27 & 83.69 & 97.26 & 94.59 & 95.86 & 83.95 & \textcolor{red}{87.96} & 85.71 & 96.47 & 96.95 & 96.69 \\  
        \multirow{1}{*}{TSAR-MVS + MP.} & 89.67 & \textcolor{red}{84.39} & \textbf{86.88} & \textcolor{red}{98.15} & \textcolor{red}{96.50} & \textbf{97.31} & 88.14 & \textbf{88.11} & \textbf{88.02} & 97.42 & \textbf{97.44} & \textbf{97.42} \\  
        \hline
        \end{tabular}%
    }
    \captionsetup{labelfont={color=black}}
  \caption{\textcolor{black}{Accuracy, completeness and F$_1$ score of point clouds on ETH3D dataset at threshold $2cm$ and $10cm$.}}
  \label{table:ETH3D}%
\end{table}%

\begin{table}
  \centering
  \renewcommand{\arraystretch}{1.05} 
    \resizebox{0.8\linewidth}{!}{
        \begin{tabular}{c|ccc|ccc|ccc|ccc}
        \hline
        \multirow{3}{*}{Method} & \multicolumn{6}{c|}{Fountain} & \multicolumn{6}{c}{HerzJesu} \\
        \cline{2-13} & \multicolumn{3}{c|}{$2cm$} & \multicolumn{3}{c|}{$10cm$} & \multicolumn{3}{c|}{$2cm$} & \multicolumn{3}{c}{$10cm$}  \\
        \cline{2-13} & Acc. & Comp. & F$_1$ & Acc. & Comp. & F$_1$ & Acc. & Comp. & F$_1$ & Acc. & Comp. & F$_1$ \\   
        \hline 
        \multirow{1}{*}{COLMAP\cite{COLMAP}} & 89.74 & 55.87 & 68.87 & 98.91 & 73.73 & 84.48 & 85.21 & 46.14 & 59.86 & 97.34 & 55.63 & 70.8 \\
        \multirow{1}{*}{OpenMVS\cite{openmvs}} & 79.62 & 70.47 & 74.77 & 91.63 & 83.49 & 87.37 & 79.76 & 61.85 & 69.67 & 92.53 & 70.37 & 79.94 \\
        \multirow{1}{*}{IterMVS-LS\cite{Iter-MVS}} & 83.02 & 69.45 & 75.63 & 93.65 & 82.61 & 87.78 & 82.97 & 60.14 & 69.73 & 94.55 & 69.83 & 80.33 \\
        \multirow{1}{*}{ACMM\cite{ACMM}} & 85.89 & 67.32 & 75.48 & 94.23 & 81.29 & 87.28 & 83.78 & 57.56 & 68.24 & 94.86 & 67.85 & 79.11 \\
        \multirow{1}{*}{ACMP\cite{ACMP}} & 86.26 & 68.53 & 76.38 & 94.55 & 82.15 & 87.91 & 84.35 & 59.73 & 69.94 & 95.19 & 69.72 & 80.49 \\
        \multirow{1}{*}{ACMMP\cite{ACMMP}} & \textbf{88.53} & 73.75 &  \textcolor{red}{80.47} & \textbf{96.18} & 84.36 & 89.88 & \textbf{86.38} & 63.41 & 73.13 & \textbf{96.81} & 72.65 & 83.01 \\
        \hline 
        \multirow{1}{*}{TSAR-MVS + GP.} & 82.27 & \textcolor{red}{78.32} & 80.25 & 93.34 & \textcolor{red}{87.32} & \textcolor{red}{90.23} & 82.27 & \textcolor{red}{66.82} & \textcolor{red}{73.74} & 94.3 & \textcolor{red}{74.69} & \textcolor{red}{83.36} \\
        \multirow{1}{*}{TSAR-MVS + MP.} & \textcolor{red}{87.65} & \textbf{79.29} & \textbf{83.26} & \textcolor{red}{95.69} & \textbf{87.51} & \textbf{91.42} & \textcolor{red}{85.43} & \textbf{68.28} & \textbf{75.90} & \textcolor{red}{96.47} & \textbf{75.38} & \textbf{84.63}  \\
        \hline 
        \end{tabular}%
    }
    \captionsetup{labelfont={color=black}}
  \caption{\textcolor{black}{Accuracy, completeness and F$_1$ score of point clouds on partial scenes of Strecha dataset (\emph{Fountain} and \emph{HerzJesu}) at threshold $2cm$ and $10cm$.}}
  \label{table:strecha}%
\end{table}%

\begin{table}
  \centering
  \renewcommand{\arraystretch}{1.05} 
    \resizebox{0.46\linewidth}{!}{
        \begin{tabular}{c|ccc|ccc}
        \hline
        \multirow{2}{*}{Method} & \multicolumn{3}{c|}{Intermediate} & \multicolumn{3}{c}{Advanced} \\
        \cline{2-7} & Pre. & Rec. & F$_1$ & Pre. & Rec. & F$_1$ \\   
        \hline 
        \multirow{1}{*}{COLMAP\cite{COLMAP}} & 43.16 & 44.48 & 42.14 & 31.57 & 23.96 & 27.24 \\  
        \multirow{1}{*}{OpenMVS\cite{openmvs}} & 46.01 & 72.38 & 55.11 & 33.77 & 39.40 & 34.43 \\ 
        \multirow{1}{*}{IterMVS-LS\cite{Iter-MVS}} & 47.53 & 73.69 & 56.94 & 28.70 & 44.19 & 34.17 \\  
        \multirow{1}{*}{ACMM\cite{ACMM}} & 49.19 & 70.85 & 57.27 & \textcolor{red}{35.63} & 34.90 & 34.02 \\ 
        \multirow{1}{*}{ACMP\cite{ACMP}} & 49.06 & 73.58 & 58.41 & 34.57 & 42.48 & 37.44 \\  
        \multirow{1}{*}{ACMMP\cite{ACMMP}} & \textbf{53.28} & 68.50 & 59.38 & 33.79 & 44.64 & 37.84 \\ 
        \hline 
        \multirow{1}{*}{PCF-MVS\cite{PCF-MVS}} & 50.04 & 58.85 & 53.39 & \textbf{35.84} & 34.35 & 34.59 \\  
        \multirow{1}{*}{SMU-MVSNet\cite{SMU-MVSNet}} & 39.67 & 63.41 & 47.81 & --- & --- & --- \\
        \multirow{1}{*}{AGG-CVCNet\cite{AGG-CVCNet}} & 49.04 & 71.71 & 57.81 & 35.33 & 28.28 & 28.96 \\
        \hline 
        \multirow{1}{*}{TSAR-MVS + GP.} & 52.03 & \textcolor{red}{73.83} & \textcolor{red}{59.94} & 34.56 & \textcolor{red}{47.28} &  \textcolor{red}{37.87} \\  
        \multirow{1}{*}{TSAR-MVS + MP.} & \textcolor{red}{53.15} & \textbf{75.52} & \textbf{62.10} & 33.85 & \textbf{48.75} & \textbf{38.63} \\  
        \hline
        \end{tabular}%
    }
    \captionsetup{labelfont={color=black}}
  \caption{\textcolor{black}{Precision, recall and F$_1$ score of point clouds on Tanks \& Temples dataset at given threshold.}}
  \label{table:TNT}%
\end{table}%

In order to achieve faster inference times during the actual experiments, we downsampled the images to half of the original resolution in both the ETH3D benchmark and the Strecha dataset, whereas adopting the original resolution in the TNT dataset. This adaptation does not compromise the robustness of our approach while enhancing its efficiency.

Our method is implemented on a system equipped with an Intel(R) Core(TM) i7-10700 CPU @ 2.90GHz and an NVIDIA GeForce RTX 3080 graphics card. 
Concerning the parameter selection, in the joint hypothesis filter, $\{\eta, \tau\}=\{0.8, 0.7\}$. 
Concurrently, in the iterative correlation refinement, $\{\sigma, \alpha, \beta, \delta, \mu\}=\{0.3, 2, 3, 5, 24\}$.
Lastly, in the textureless-aware segmentation,  $\{\varepsilon, \upsilon, \kappa, \gamma\}=\{5, 15, 0.8, 0.1\}$. 
We employ a matching strategy that uses every other row and column to speed up the computation of the NCC \cite{COLMAP}. 

\subsection{Point Cloud Evaluation}
To evaluate the effectiveness of our method, we compare the accuracy (Acc.), completeness (Comp.), and F$_1$ score of our reconstructed point clouds with other methods. 
\textcolor{black}{Specifically, we compare our work against well-known traditional MVS methods including COLMAP~\cite{COLMAP}, OpenMVS~\cite{openmvs}, IterMVS-LS~\cite{Iter-MVS}, ACMM~\cite{ACMM}, ACMP~\cite{ACMP}, ACMMP~\cite{ACMMP}, APD-MVS~\cite{APD-MVS} and partial closely related methods including TAPA-MVS~\cite{TAPA-MVS}, PCF-MVS~\cite{PCF-MVS}, MG-MVS~\cite{MG-MVS}, SMU-MVSNet~\cite{SMU-MVSNet} and AGG-CVCNet~\cite{AGG-CVCNet}. 
The corresponding quantitative results of the ETH3D dataset, the Strecha dataset and the TNT dataset are respectively shown in Tab. \ref{table:ETH3D}, Tab. \ref{table:strecha} and Tab. \ref{table:TNT}. Note that the best results are marked in bold while the second-best results are marked in red.}
Since both the ETH3D dataset and the TNT dataset provides a benchmark, we directly copied the quantitative results of other methods from their official websites.
Differently, since Strecha only provides the point cloud ground truth, we achieved the quantitative results of other methods by replicating them by ourselves.

\textcolor{black}{Regarding the baseline for our method, TSAR-MVS + GP. is gengrated by adopting Gipuma~\cite{Gipuma} with adaptive checkerboard propagation~\cite{ACMM} as our baseline, for it offers the best performance and time trade-off. To further validate the performance improvements, we generate TSAR-MVS + MP. by adopting ACMMP~\cite{ACMMP} as the new baseline of our method.}

\begin{figure*}
\centering
\includegraphics[width=\linewidth]{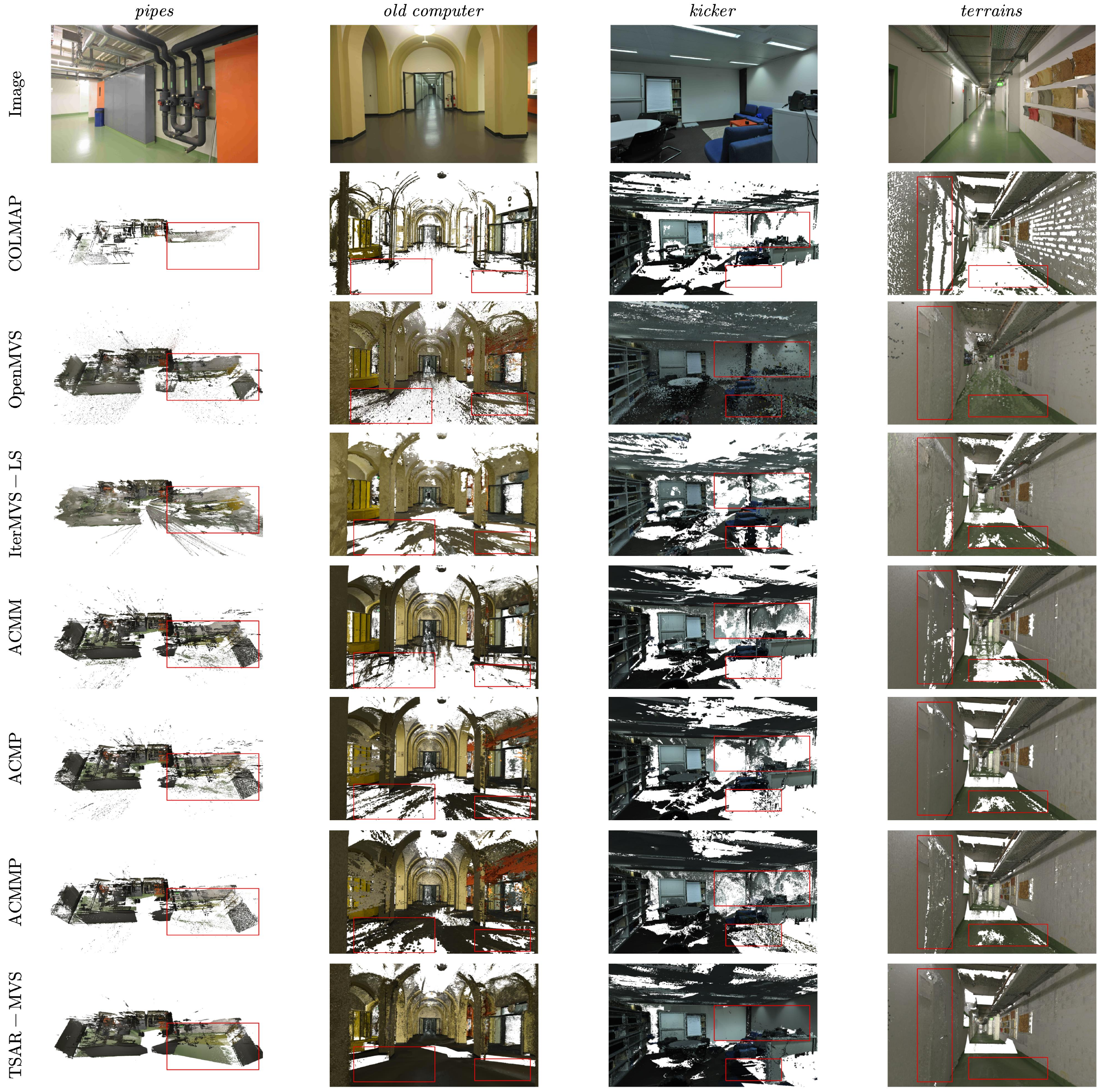}
\caption{Visualized point cloud results between different methods on partial scenes of ETH3D datasets (\emph{pipes}, \emph{old computer}, \emph{kicker} and \emph{terrains}). TSAR-MVS can effectively reconstruct textureless areas such as floors and walls without detail distortion. }
\label{fig7}
\end{figure*}

\begin{figure*}
\centering
\includegraphics[width=\linewidth]{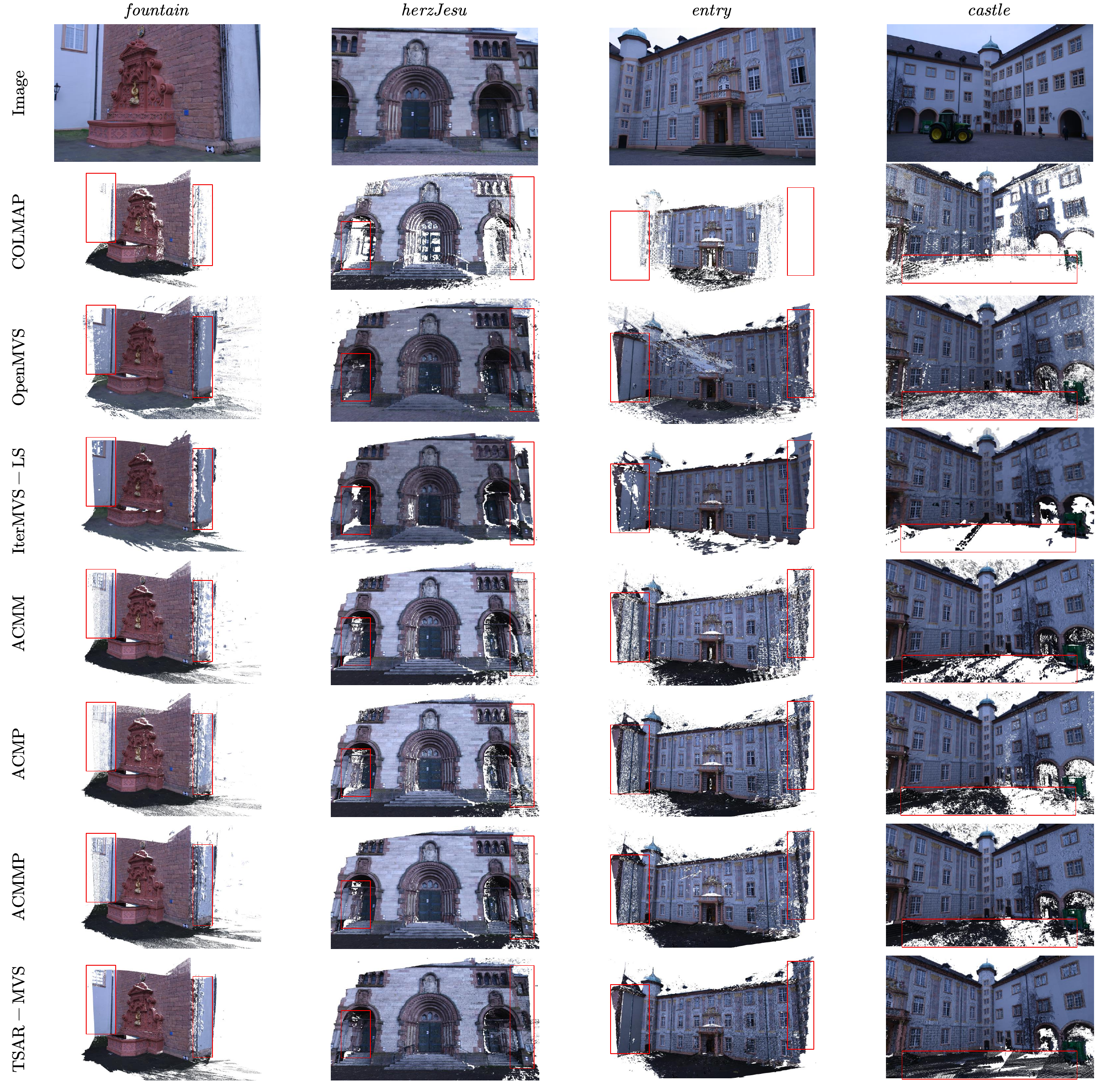}
\caption{Visualized point cloud results between different methods on partial scenes of Strecha datasets (\emph{fountain}, \emph{herzJesu}, \emph{entry} and \emph{castle}).}
\label{fig8}
\end{figure*}

\begin{figure*}
\centering
\includegraphics[width=\linewidth]{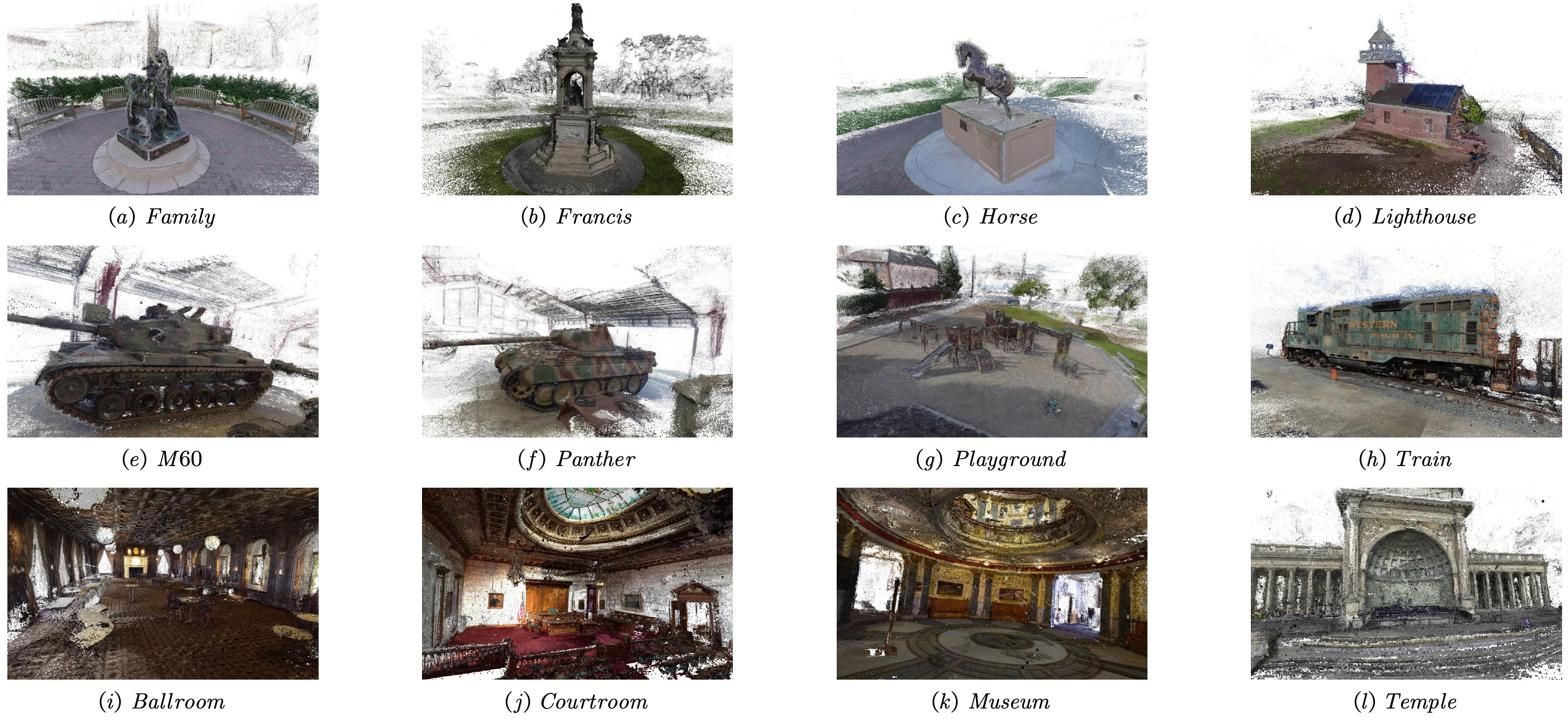}
\captionsetup{labelfont={color=black}}
\caption{\textcolor{black}{Reconstructed point clouds of our method on Tanks \& Temples dataset without any fine-tuning.}}
\label{fig_TNT}
\end{figure*}

\textcolor{black}{Considering the ETH3D dataset, although TSAR-MVS + GP. adopts a modified version of Gipuma as our baseline, it achieves significant improvements in results, which is comparable to ACMMP. Moreover, TSAR-MVS + MP. achieves the highest $F_1$ score among all methods, not only significantly outperforms ACMMP and but also exceeds current state-of-the-art methods APD-MVS, validating the effectiveness of our approach. 
Regarding the TNT dataset and the Strecha dataset, both TSAR-MVS + GP. and TSAR-MVS + MP. distinctly surpasses other approaches, manifesting superiority in both completeness and overall quality metrics, indicating the resilient competitiveness inherent in our approach and its capacity to match stride with the most advanced solutions.}

Qualitative results from the ETH3D dataset, the TNT dataset and the Strecha dataset are respectively illustrated in Fig. \ref{fig7}, Fig. \ref{fig8} and Fig. \ref{fig_TNT}, . Clearly, our method yields the most comprehensive results, especially in textureless planar areas like floors and walls, without inducing any conspicuous detail distortion. This observation demonstrates the texture-adaptive capabilities of our proposed method, allowing for the reconstruction of textureless areas while preserving fine details. More qualitative results are available in the supplementary material. 

\begin{figure}
\centering
\includegraphics[width=0.95\linewidth]{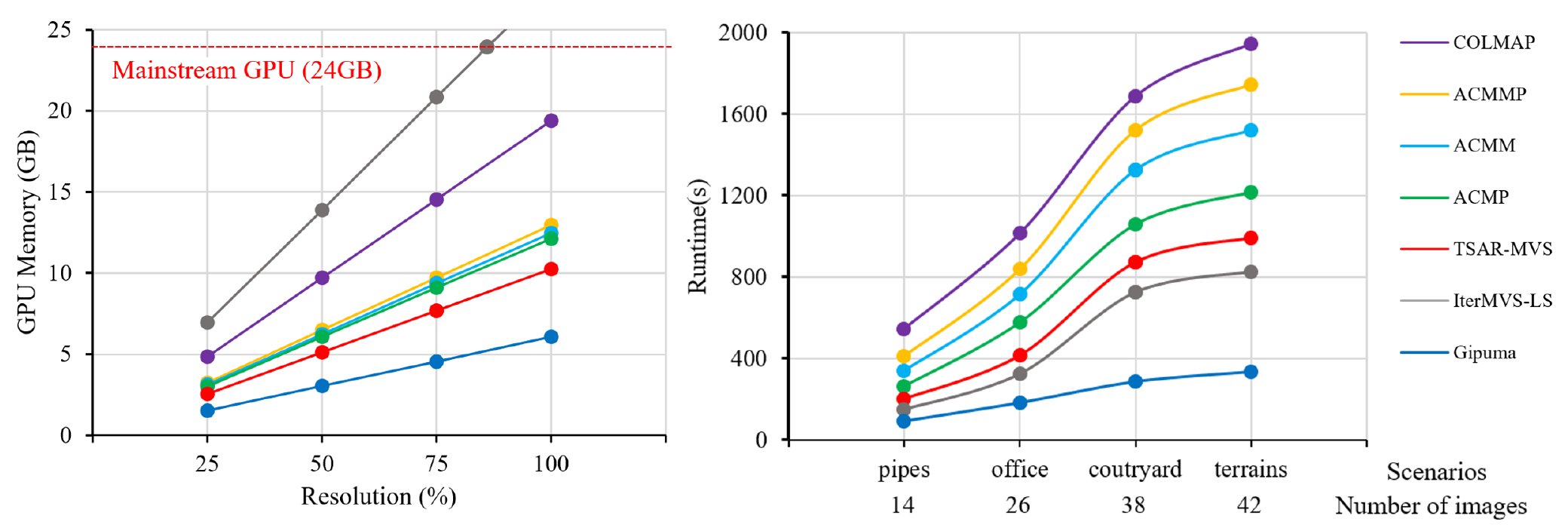}
\caption{Total runtime (second) and GPU memory (GB) between different methods before fusion on ETH3D dataset.  
When comparing the runtime, we select partial scenarios (\emph{pipes}, \emph{office}, \emph{courtyard} and \emph{terrains}) and scale their image resolution down to 50\%. 
Conversely, when evaluating memory consumption, we align the number of source images to 10 across all methods and scale them to different resolutions.}
\label{fig9}
\end{figure}

It's also worth noting that our method offers a significant advantage in terms of computational efficiency. Fig. \ref{fig9} presents the runtime and GPU memory overhead between different methods. To exclude the influence of unrelated variables, all methods were implemented on the same system, with the hardware configuration provided in Section \ref{section:Dataset and Settings}.
Despite IterMVS-LS and Gipuma are faster than TSAR-MVS, the GPU memory overhead demanded by IterMVS-LS is three times that of TSAR-MVS,  and Gipuma's reconstruction quality significantly falls short of TSAR-MVS.
Furthermore, TSAR-MVS exhibits superior performance compared to COLMAP, ACMM, ACMP, and ACMMP in terms of both time and GPU memory overhead.
This proves that our method is not only effective but also highly practical for applications that require both precision and speed.

\subsection{Analysis Experiments}
\subsubsection{Ablation Study}
To validate the design choices in our method, we conducted a series of experiments to corroborate the effectiveness of each constituent component. 
Tab. \ref{ablation} exhibits the percentage of unduplicated fused pixels with absolute depth errors less than $2cm$ and $10cm$ of 13 scenes in ETH3D training dataset. 
Here we term unduplicated fused pixels as pixels that can be fused by at least two views without repetition. 
For any 3D point, its accuracy is consistent across depth maps from multiple viewpoints. Therefore, to avoid the impact of repetitive results, for each 3D point, we selected one pixel from all the viewpoints used for fusion to assess its accuracy, and discarding pixels from the rest of the viewpoints. 

\textcolor{black}{In Tab. \ref{ablation}, row 1-2 respectively presents the results derived from Gipuma~\cite{Gipuma} plus adaptive checkerboard propagation and ACMM~\cite{ACMM}.  
Row 10 provides the results derived from our method with modified Gipuma as prior input.
Rows 3-9 respectively illustrate the results when the joint hypothesis filtering is omitted (w/o JHF), the disparity discontinuity is deleted (w/o DD), the confidence estimator is taken out (w/o CE), the iterative correlation refinement is excluded (w/o ICR), the superpixel planarization is removed (w/o SP), the WMF is discarded (w/o WMF) and the textureless-aware segmentation is erased (w/o TS). }

\begin{table*}
  \centering
  \renewcommand{\arraystretch}{1.05}
  \resizebox{0.85\textwidth}{!}{
    \begin{tabular}{c|c|c|ccccccc|cccccc}
    \hline
    \multirow{2}{*}{error} & \multirow{2}{*}{method} & \multirow{2}{*}{Ave.} & \multicolumn{7}{c|}{indoor}                           & \multicolumn{6}{c}{outdoor} \\
    \cline{4-16}          &       &       & delive. & kicker & office & pipes & relief & relief\_2 & terrace & courty. & electro & façade & meadow & playgr. & terrace \\
    \hline
    \multirow{10}{*}{$2cm$}
    & Prior & 21.99 & 19.47 & 22.55 & 17.23 & 22.97 & 27.51 & 26.78 & 23.04 & 22.41 & 29.05 & 19.29 & 14.67 & 17.62 & 23.3 \\ 
    & ACMM & 32.58 & 27.64 & 32.6  & 29.13 & 37.09 & 38.63 & 35.45 & 29.58 & 33.59 & 40.26 & 26.34 & \textbf{35.43} & 26.43 & 31.43 \\
    \cdashline{2-16}
    & w/o JHF & 41.45 & 35.93 & 31.49 & 29.96 & 45.57 & 44.26 & 46.95 & 50.36 & 46.41 & 53.73 & 34.07 & 23.27 & 46.63 & 50.16 \\
    & w/o DD & 41.98 & 36.99 & 31.9 & 32.08 & 45.93 & 44.03 & 46.77 & 50.97 & 46.69 & 54.26 & 34.13 & 24.65 & 46.85 & 50.43 \\
    & w/o CE & 42.19 & 36.37 & 32.12 & 34.36 & 45.77 & 44.1 & 46.76 & 51.23 & 46.88 & 54.51 & 34.25 & 24.73 & 47.12 & 50.27 \\
    \cdashline{2-16}
    & w/o ICR & 36.76 & 33.28 & 24.64 & 18.31 & 31.53 & 41.97 & 43.83 & 47.88 & 45.23 & 52.32 & 32.99 & 15.91 & 41.91 & 48.09 \\
    & w/o SP & 39.42 & 34.53 & 28.14 & 27.4 & 37.82 & 42.64 & 44.63 & 49.35 & 46.27 & 53.18 & 33.46 & 22.48 & 43.84 & 48.76 \\
    & w/o WMF  & 40.87 & 35.26 & 30.52 & 33.65 & 41.27 & 42.78 & 44.85 & 49.12 & 46.82 & 53.98 & 33.71 & 25.35 & 44.37 & 49.57 \\
    \cdashline{2-16}
    & w/o TS & 40.15 & 34.38 & 28.34 & 25.86 & 42.31 & 44.58 & 47.31 & 52.1  & 47.18 & 53.25 & 33.41 & 20.12 & 43.68 & 49.37 \\
    & TSAR-MVS & \textbf{44.76} & \textbf{37.34} & \textbf{35.71} & \textbf{44.76} & \textbf{46.54} & \textbf{44.82} & \textbf{47.52} & \textbf{53.33} & \textbf{48.4}  & \textbf{55.15} & \textbf{34.98} & 33.28 & \textbf{48.95} & \textbf{51.11} \\
    \hline
    \multirow{10}{*}{$10cm$} 
    & Prior & 24.67 & 21.01 & 26.47 & 20.35 & 27.67 & 27.93 & 27.27 & 24.51 & 26.14 & 32.29 & 24.89 & 17.69 & 19.92 & 24.59 \\
    & ACMM & 37.11 & 33.91 & 39.21 & 34.55 & 39.6  & 39.66 & 36.54 & 31.84 & 37.95 & 45.21 & 35.2  & 42.07 & 30.45 & 36.19 \\
    \cdashline{2-16}
    & w/o JHF & 48.78 & 42.93 & 38.64 & 35.03 & 50.73 & 47.33 & 49.77 & 57.85 & 53.57 & 61.17 & 48.93 & 36.1  & 53.64 & 58.45 \\
    & w/o DD  & 49.31 & 44.16 & 39.08 & 36.94 & 51.53 & 46.97 & 49.6 & 58.04 & 53.85 & 61.44 & 48.93 & 37.96 & 53.91 & 58.61 \\
    & w/o CE  & 49.52 & 43.34 & 39.2 & 39.46 & 51.46 & 47.05 & 49.68 & 58.12 & 54.03 & 61.6 & 48.94 & 38.12 & 54.18 & 58.55 \\
    \cdashline{2-16}
    & w/o ICR & 43.38 & 38.64 & 28.89 & 25.66 & 35.77 & 44.02 & 46.48 & 54.41 & 52    & 58.83 & 46.85 & 28.23 & 48.33 & 55.81 \\
    & w/o SP & 46.30 & 40.52 & 34.18 & 34.39 & 41.48 & 44.86 & 47.42 & 55.89 & 53.21 & 59.74 & 47.54 & 35.37 & 50.63 & 56.65 \\
    & w/o WMF & 47.88 & 41.73 & 37.42 & 40.71 & 45.05 & 45.07 & 47.67 & 55.63 & 53.87 & 60.57 & 47.83 & 38.06 & 51.34 & 57.53 \\
    \cdashline{2-16}
    & w/o TS & 47.14 & 40.18 & 35.23 & 31.42 & 47.82 & 47.45 & 50.37 & 55.44 & 54.78 & 59.49 & 48.34 & 33.18 & 51.9  & 57.17 \\
    & TSAR-MVS & \textbf{52.40} & \textbf{45.52} & \textbf{43.14} & \textbf{52.56} & \textbf{52.17} & \textbf{47.58} & \textbf{50.51} & \textbf{58.6}  & \textbf{56.1}  & \textbf{62.14} & \textbf{49.98} & \textbf{47.3}  & \textbf{56.25} & \textbf{59.32} \\
    \hline
    \end{tabular}%
    }
  \caption{Quantitative results of the ablation studies on percentage of unduplicated fused pixels at depth tolerance $2cm$ and $10cm$ of ETH3D training dataset.}
  \label{ablation}%
\end{table*}%

\begin{figure*}
\centering
\includegraphics[width=\linewidth]{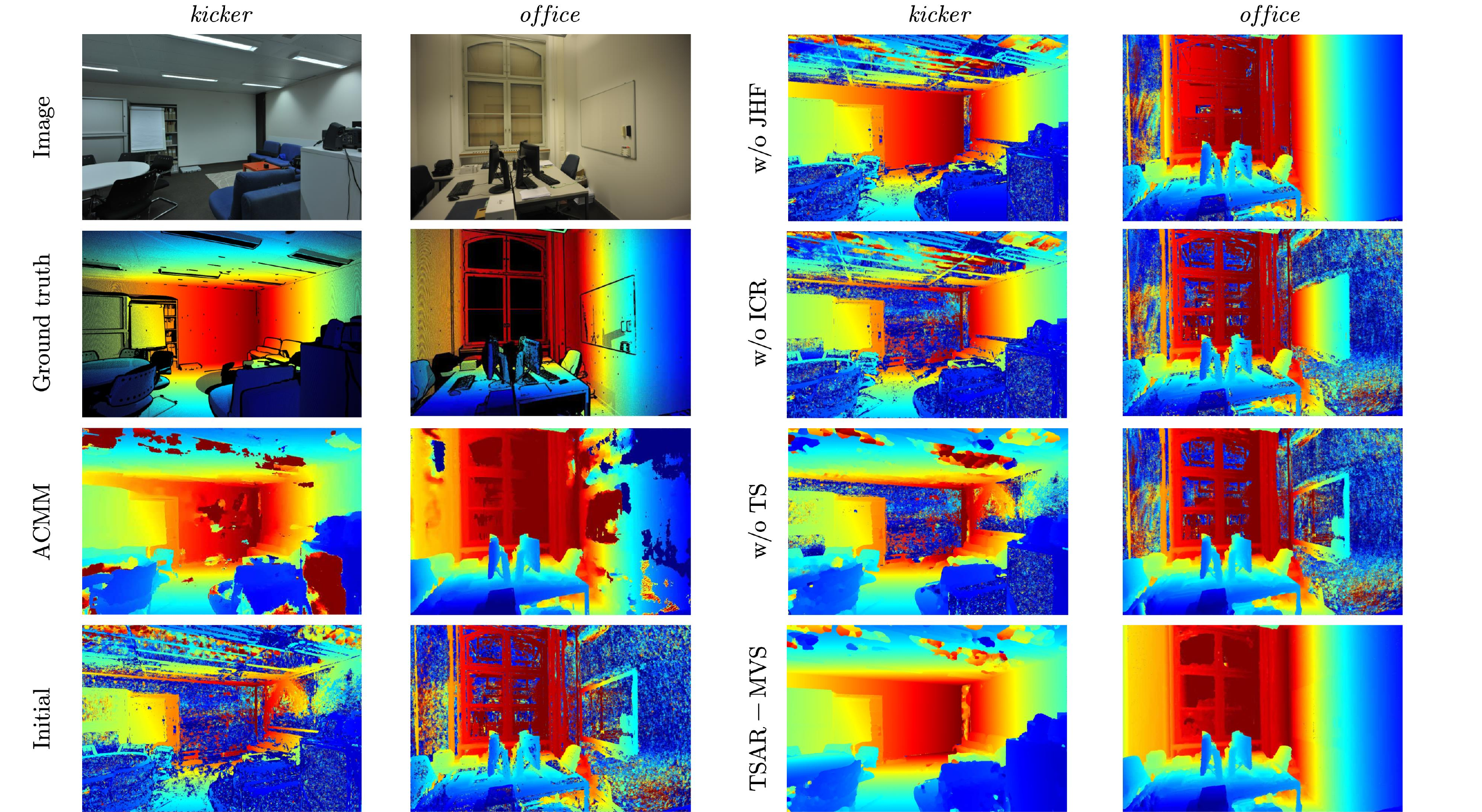}
\caption{Visualized depth maps by various ablation methods on partial scenes of ETH3D datasets (\emph{kicker} and \emph{office}). Where the heat value is higher, the larger its depth will be, and vice versa. Black areas in (b) have no ground truth data. TSAR-MVS shows remarkable strength in reconstructing textureless areas. }
\label{fig6}
\end{figure*}

\textcolor{black}{As can be seen from Tab. \ref{ablation}, each proposed component contributes significantly to our pipeline.
Specifically, w/o ICR is worse than w/o JHF and w/o TS. Hence ICR contributes more than both JHF and TS, further emphasizing its significance. 
Moreover, although w/o JHF and w/o TS achieve similar result, w/o JHF outperforms w/o TS in scenarios with more textureless areas such as \emph{pipes} and \emph{office}, since TS is targeted towards handling textureless areas, whereas JHF is beneficial for retaining fine detail. This observation is further substantiated in Fig. \ref{fig6}. Furthermore, w/o DD essentially aligns with w/o CE and w/o WMF slightly surpasses w/o SP, evidencing their respective contributions within each submodule.}

\textcolor{black}{Furthermore, Fig. \ref{fig6} provide depth images displayed in pseudo colors for intuitive comparison. 
Without filtering, w/o JHF may erroneously estimate partial superpixels and causing severe deviations during planarization. 
Without refinement, w/o ICR not only fails in reconstructing textureless areas but also causing detail distortion. 
Without segmentation, w/o TS cannot reconstruct any lager textureless areas. 
In contrast, the TSAR-MVS produces the most realistic results, avoiding both plane shifting and detail distortion.}

\begin{table*}
  \centering
  \renewcommand{\arraystretch}{1.05}
  \resizebox{0.85\textwidth}{!}{
    \begin{tabular}{cccccc|cccccc|cccccc}
    \hline
    \multicolumn{6}{c}{Joint Hypothesis Filtering} & \multicolumn{6}{|c|}{Iterative Correlation Refinement} & \multicolumn{6}{c}{Textureless-aware Segmentation} \\
    \hline
    \multicolumn{1}{c|}{$\eta$} & 0.4 & 0.6 & \textcolor{blue}{0.8} & 1.0 & 1.2 & \multicolumn{1}{c|}{$N_t$} & 1 & 2 & 3 & \textcolor{blue}{4} & 5 & \multicolumn{1}{c|}{$\kappa$} & 0.3 & \textcolor{blue}{0.8} & 1.3 & 2.0 & 3.0 \\ 
    \cdashline{1-18}
    \multicolumn{1}{c|}{F$_1$} & 82.85 & \textcolor{red}{83.43} & \textbf{83.69} & 83.35 & 82.67 & \multicolumn{1}{c|}{F$_1$} & 81.25 & 82.54 & 83.36 & \textcolor{red}{83.69} & \textbf{83.87} & \multicolumn{1}{c|}{F$_1$} & \textbf{83.78} & \textcolor{red}{83.69} & 83.27 & 82.74 & 81.49 \\

    \hline
    \multicolumn{1}{c|}{$\tau$} & 0.1 & 0.3 & 0.5 & \textcolor{blue}{0.7} & 0.9 & \multicolumn{1}{c|}{$\mu$} & 4 & 8 & 16 & \textcolor{blue}{24} & 32 & \multicolumn{1}{c|}{$\gamma$} & 0.05 & \textcolor{blue}{0.1} & 0.2 & 0.3 & 0.5\\ 
    \cdashline{1-18}
    \multicolumn{1}{c|}{F$_1$} & 82.67 & 83.12 & \textcolor{red}{83.44} & \textbf{83.69} & 82.45 & \multicolumn{1}{c|}{F$_1$} & 82.74 & 83.04 & 83.37 & \textbf{83.69} &  \textcolor{red}{83.46} & \multicolumn{1}{c|}{F$_1$} & \textcolor{red}{83.46} & \textbf{83.69} & 83.24 & 82.86 & 82.37 \\

    \hline
    \multicolumn{6}{c}{Superpixel} & \multicolumn{6}{|c|}{Weighted Median Filtering} & \multicolumn{6}{c}{Hough Line Detection} \\
    \hline
    \multicolumn{1}{c|}{$\sigma$} & 0.1 & 0.2 & \textcolor{blue}{0.3} & 0.4 & 0.5 & \multicolumn{1}{c|}{$l_A$} & $3 \cdot 2^{(N_t - N_c)}$ & 3 & \textcolor{blue}{$5 \cdot 2^{(N_t - N_c)}$} & 5 & $7 \cdot 2^{(N_t - N_c)}$ & \multicolumn{1}{c|}{$\varepsilon$} & 2 & \textcolor{blue}{5} & 10 & 15 & 20 \\ 
    \cdashline{1-18}
    \multicolumn{1}{c|}{F$_1$} & 83.13 & \textcolor{red}{83.47} & \textbf{83.69} & 82.95 & 82.27 & \multicolumn{1}{c|}{F$_1$} & 83.39 & 82.63 & \textbf{83.69} & 82.95 & 83.75 & \multicolumn{1}{c|}{F$_1$} & \textbf{83.75} & \textcolor{red}{83.69} & 83.23 & 82.87 & 82.43 \\
    \hline

    \hline
    \multicolumn{1}{c|}{$S_{sp}$} & 100 & 200 & \textcolor{blue}{400} & 800 & 1600 & \multicolumn{1}{c|}{$\delta$} & 1 & 3 & \textcolor{blue}{5} & 7 & 9 & \multicolumn{1}{c|}{$\upsilon$} & 5 & 10 & \textcolor{blue}{15} & 20 & 25 \\ 
    \cdashline{1-18}
    \multicolumn{1}{c|}{F$_1$} & 82.78 & \textcolor{red}{83.30} & \textbf{83.69} & 83.05 & 82.16 & \multicolumn{1}{c|}{F$_1$} & 82.48 & 83.25 & 83.69 & \textcolor{red}{83.80} & \textbf{83.85} & \multicolumn{1}{c|}{F$_1$} & 83.25 & \textcolor{red}{83.48} & \textbf{83.69} & 83.31 & 82.80 \\
    \hline
    \end{tabular}%
    }
  \captionsetup{labelfont={color=black}}
  \caption{\textcolor{black}{Quantitative evaluation of hyperparameter variability on the F$_1$ Score at threshold $2cm$ of ETH3D training dataset.}}
  \label{hyper}%
\end{table*}%

\textcolor{black}{To justify our hyperparameter choices, we conducted ablation studies on key parameters, as shown in Table \ref{hyper}. The best, second-best, and chosen parameters are highlighted in bold, red, and blue, respectively. As can be seen, apart from significant parameters such as $N_t$, $\kappa$ and $S_{sp}$, most parameters exhibit low sensitivity. 
Concerning filtering, $\eta$ and $\tau$ demonstrate that our parameter selection is made towards achieving the optimal results.
Regarding refinement, the result continuously improves as $N_t$ increases, yet we choose $N_t = 4$ to trade-off between time and performance. 
This reasoning also underlies the selection of values for $\kappa$ in segmentation, $\delta$ in WMF, and $\varepsilon$ in line detection. 
By comparing $\kappa$ and $\gamma$ in segmentation, we found significant sensitivity variations across different parameters. 
Finally, $S_{sp}$ in superpixel and $l_A$ in WMF respectively indicates the significance of appropriate size for superpixels and adaptive window size for filtering.}

\subsubsection{Generalization Performance}
\textcolor{black}{Moreover, to demonstrate the generalization capability of our method, we conducted a series of experiments by respectively substituting the prior input into modified Gipuma~\cite{Gipuma}, COLMAP~\cite{COLMAP} ACMM~\cite{ACMM}, and ACMMP~\cite{ACMMP}, as shown in Tab. \ref{generalization}. }
We adopt the Gipuma with adaptive checkerboard propagation ~\cite{ACMM} as our first baseline, thereby pursuing high-quality reconstruction while striving for temporal efficiency.
While Gipuma excels in speed, its reconstruction results leave much to be desired. Although ACMM performs better, its multi-scale framework tends to be time-consuming. Hence, by incorporating adaptive checkerboard propagation into Gipuma, we can ensure both speed and the provision of excellent depth priors for reconstruction, creating a beneficial blend of both models.

As indicated by the results presented in Table \ref{generalization}, the integration of our method with each of these initial depth inputs engenders noticeable improvements. 
Comparatively, the results achieved by using COLMAP as baseline demonstrated greater performance gains than those acquired using ACMM. 
This suggests that our method yields more significant enhancements when faced with unreliable initial depth. 
Given that ACMM exhibits superior performance than COLMAP in addressing textureless area challenges, it further validates that our method is especially proficient at refining depth when initial credible depth values are notably absent within textureless regions.
These findings collectively corroborate the potential of our TSAR-MVS to be effectively applied across various techniques, thereby demonstrating its strong generalization capability.

\begin{table}
  \centering
  \renewcommand{\arraystretch}{1.05} 
    \resizebox{0.9\linewidth}{!}{
        \begin{tabular}{c|ccc|ccc|c|ccc|ccc|c}
        \hline
        \multirow{3}{*}{Method} & \multicolumn{7}{c|}{$2cm$} & \multicolumn{7}{c}{$10cm$} \\
        \cline{2-15} & \multicolumn{3}{c|}{w/o. TSAR-MVS} & \multicolumn{3}{c|}{w/. TSAR-MVS} & \multicolumn{1}{c|}{Change} & \multicolumn{3}{c|}{w/o. TSAR-MVS} & \multicolumn{3}{c|}{w/. TSAR-MVS} & \multicolumn{1}{c}{Change}  \\
        \cline{2-15} & Acc. & Comp. & F$_1$ & Acc. & Comp. & F$_1$ & F$_1$ & Acc. & Comp. & F$_1$ & Acc. & Comp. & F$_1$ & F$_1$  \\   
        \hline 
        \multirow{1}{*}{COLMAP\cite{COLMAP}} & 91.85 & 55.13 & 67.66 & 85.51 & 73.09 & 78.34 & \textcolor{red}{10.68 $\uparrow$} & 98.75 & 79.47 & 87.61 & 96.05 & 89.59 & 92.39 & \textcolor{red}{4.78 $\uparrow$} \\
        \multirow{1}{*}{Gipuma++\cite{Gipuma}} & 87.65 & 66.53 & 75.36 & 85.32 & 82.27 & 83.69 & \textcolor{red}{8.33 $\uparrow$} & 96.64 & 85.73 & 91.21 & 97.26 & 94.59 & 95.86 & \textcolor{red}{4.65 $\uparrow$} \\
        \multirow{1}{*}{ACMM\cite{ACMM}} & 90.67 & 70.42 & 78.86 & 89.14 & 81.59 & 85.10 & \textcolor{red}{6.24 $\uparrow$} & 98.12 & 86.40 & 91.70 & 97.54 & 94.99 & 96.19 & \textcolor{red}{4.49 $\uparrow$} \\
        \multirow{1}{*}{ACMMP\cite{ACMMP}} & 90.63 & 77.61 & 83.42 & 89.67 & 84.39 & 86.88 & \textcolor{red}{3.46 $\uparrow$} & 97.99 & 93.32 & 95.54 & 98.15 & 96.50 & 97.64 & \textcolor{red}{2.1 $\uparrow$} \\
        \hline
        \end{tabular}%
    }
    \captionsetup{labelfont={color=black}}
    \caption{\textcolor{black}{Increase of Accuracy, completeness and F$_1$ score of point clouds on ETH3D training dataset at threshold $2cm$ and $10cm$.}}
    \label{generalization}%
\end{table}%

\section{Conclusion}

In this paper, we introduced TSAR-MVS, a novel multi-view stereo method explicitly designed for the reconstruction of textureless areas. 
First, we present the joint hypothesis filtering, a synergistic blend of a confidence estimator and disparity discontinuity filtering, thus enabling us to provisionally discard erroneous estimates. 
Subsequently, we implement the iterative correlation refinement, iteratively utilizing RANSAC and WMF to fill in areas with robust hypotheses derived from nearby trustworthy regions. 
Moreover, we propose the textureless-aware segmentation, a distinctive fusion of edge and line detection techniques, functions as a flexible mechanism that enables adaptive recognition of textureless regions.
\textcolor{black}{Experiments on the ETH3D, Tanks \& Temples and Strecha datasets have demonstrated that our method can significantly outperforms others and exhibits robustness to textureless areas while preserving fine details. 
Ablation studies also validate the effectiveness of each proposed component and demonstrate the strong generalization capability of our method.}

\textcolor{black}{However, our proposed method primarily addresses textureless areas through leveraging spatial neighborhood information, without considering multi-view geometric constraint. 
}
\textcolor{black}{
Therefore, our method struggles in handling reflective surfaces, frequently misidentify highlight areas as well-textured areas, thus causing reconstruction errors. The absence of superpixels with varying sizes may cause deviations during refinement. Furthermore, our method may struggle when encountering curved boundaries, leaving room for further improvement.
}
We look forward to further refining our approach and exploring its potential applications in a broader range of scenarios.

\section*{Acknowledgement}
This work was supported by National Natural Science Foundation of China (Grant No.62172392).

\bibliography{TSAR-MVS}
\bibliographystyle{elsarticle-num}

\end{document}